\documentclass[11pt,a4paper]{article}
\usepackage[hyperref]{acl2021}
\usepackage{times}
\usepackage{latexsym}

\usepackage{microtype}

\aclfinalcopy 

\usepackage{microtype}
\usepackage{graphicx}
\usepackage{subfigure}
\usepackage{booktabs} 

\usepackage{outlines}
\usepackage{babel,blindtext}
\usepackage{url}
\usepackage{tabulary}
\usepackage{tabularx}
\usepackage[normalem]{ulem}
\usepackage{enumitem}
\usepackage{multirow}
\usepackage{array}
\usepackage{makecell}
\usepackage{amsmath,amsfonts,amsthm,amssymb,amsopn,bm}
\usepackage{color,soul}
\usepackage{verbatim}
\usepackage{pifont}
\usepackage{footnote}

\newcommand\desc{\text{MDL}}
\newcommand{\hotpot}{\textsc{HotpotQA}}

\newcommand{\yes}{\texttt{yes}}
\newcommand{\no}{\texttt{no}}

\newcommand\ignore[1]{}

\newcommand{\transformer}{\ensuremath{\textsc{Transformer}}}
\newcommand{\transformerbase}{\ensuremath{\textsc{Transformer}_{\textsc{BASE}}}}

\newcommand{\longformer}{\textsc{Longformer}}
\newcommand{\longformerbase}{\ensuremath{\textsc{Longformer}_{\textsc{BASE}}}}

\newcommand{\roberta}{\textsc{RoBERTa}}
\newcommand{\distilroberta}{\ensuremath{\textsc{DistilRoBERTa}}}
\newcommand{\robertabase}{\ensuremath{\textsc{RoBERTa}_{\textsc{BASE}}}}
\newcommand{\robertalarge}{\ensuremath{\textsc{RoBERTa}_{\textsc{LARGE}}}}
\newcommand{\albert}{\textsc{ALBERT}}
\newcommand{\albertbase}{\ensuremath{\textsc{ALBERT}_{\textsc{BASE}}}}
\newcommand{\bart}{\textsc{BART}}
\newcommand{\bartbase}{\ensuremath{\textsc{BART}_{\textsc{BASE}}}}
\newcommand{\gpt}{\ensuremath{\textsc{GPT2}}}
\newcommand{\distilgpt}{\ensuremath{\textsc{DistilGPT2}}}

\title{Rissanen Data Analysis:\\Examining Dataset Characteristics via Description Length}

\author{Ethan Perez$^1$ ~~ \textbf{Douwe Kiela}$^2$ ~~ \textbf{Kyunghyun Cho}$^{1~3}$ \\
$^1$New York University, $^2$Facebook AI Research\\$^3$CIFAR Fellow in Learning in Machines \& Brains\\
  {\tt perez@nyu.edu}}

\date{}

\begin{document}
\maketitle
\begin{abstract}
We introduce a method to determine if a certain capability helps to achieve an accurate model of given data. We view labels as being generated from the inputs by a program composed of subroutines with different capabilities, and we posit that a subroutine is useful if and only if the minimal program that invokes it is shorter than the one that does not. Since minimum program length is uncomputable, we instead estimate the labels' minimum description length (MDL) as a proxy, giving us a theoretically-grounded method for analyzing dataset characteristics. We call the method Rissanen Data Analysis (RDA) after the father of MDL, and we showcase its applicability on a wide variety of settings in NLP, ranging from evaluating the utility of generating subquestions before answering a question, to analyzing the value of rationales and explanations, to investigating the importance of different parts of speech, and uncovering dataset gender bias.\footnote{Code and results at \href{https://github.com/ethanjperez/rda}{https://github.com/ethanjperez/rda}, along with a script to conduct RDA on your own dataset.}
\end{abstract}

\section{Introduction}
\label{sec:Introduction}

In many practical learning scenarios, it is useful to know what capabilities would help to achieve a good model of the data.
According to Occam's Razor, a good model is one that provides a simple explanation for the data~\citep{blumer1987occam}, which means that the capability to perform a task is helpful when it enables us to find simpler explanations of the data.
Kolmogorov complexity~\cite{kolmogorov1968three} formalizes the notion of simplicity as the length of the shortest program required to generate the labels of the data given the inputs.
In this work, we estimate the Kolmogorov complexity of the data by approximately computing the data's Minimum Description Length~\citep[MDL;][]{rissanen1978modeling}, and we examine how the data complexity changes as we add or remove different features from the input.
We name our method Rissanen Data Analysis (RDA)
after the father of the MDL principle, and we use it to examine several open questions about popular datasets, with a focus on NLP.

\begin{figure}[t]
	\centering
    \includegraphics[scale=0.6]{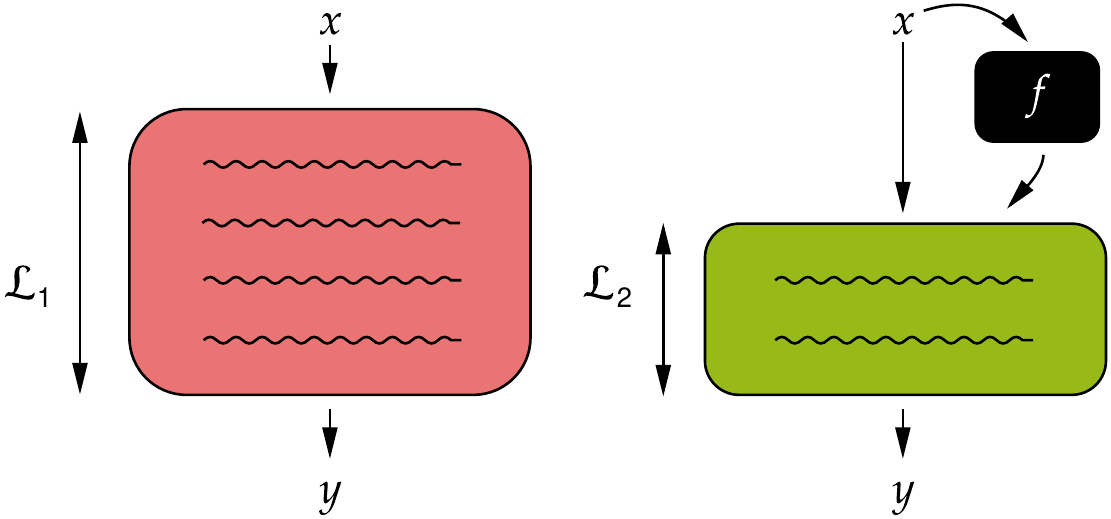}
    \caption{A capability $f$ is useful if it shortens the minimum program needed to perform a task, as measured by Minimum Description Lengths $\mathcal{L}_1$ and $\mathcal{L}_2$. For example, if $x$ is a question and $y$ is an answer, then $f$ can be an oracle that answers relevant subquestions.
    }
    \label{fig:overview}
\end{figure}

We view a capability as a function $f(x)$ that transforms $x$ in some way (e.g., adding a feature), and we say that $f$ is helpful if invoking it leads to a shorter minimum program for mapping $x$ to the corresponding label in a dataset (see Fig.~\ref{fig:overview} for an illustration).
Finding a short program is equivalent to finding a compressed version of the labels given the inputs, since the program can be run to generate the labels.
Thus, we can measure the shortest program's length by estimating the labels' maximally compressed length, or Minimum Description Length~\citep[MDL;][]{rissanen1978modeling,grunwald2004tutorial}.
While prior work in machine learning uses MDL for model optimization~\cite{hinton1993keeping}, selection~\cite{yogatama2019learning}, and model probing~\cite{voita-titov-2020-information,lovering2021predicting}, we use MDL for a very different end: to understand the data itself (``dataset probing'').

RDA addresses empirical and theoretical inadequacies of prior data analysis methods.
For example, two common approaches are to evaluate the performance of a model when the inputs are modified or ablated (1) at training and test time or (2) at test time only.
Training time input modification has been used to evaluate the usefulness of the capability to decompose a question into subquestions~\cite{min-etal-2019-multi,perez-etal-2020-unsupervised}, to access the image for image-based question-answering~\cite{antol2015vqa,zhang2016yin}, and to view the premise when detecting if it entails a hypothesis~\cite{gururangan-etal-2018-annotation,poliak-etal-2018-hypothesis,tsuchiya-2018-performance}.
However, these works evaluate performance only on held-out dev examples, a fraction of all examples in the dataset, which also are often drawn from a different distribution (e.g., in terms of quality).
To understand what datasets teach our models, we must examine the entire dataset, also giving us more examples for evaluation.
Furthermore, a capability's usefulness to a model in high-data regimes does not necessarily reflect its usefulness in low-data regimes, which have increasingly become of interest~\cite{lake2017building,guzman-etal-2019-flores,brown2020language}.
Test time ablation has been used to evaluate the capability to view word order~\cite{pham2020order,sinha2020unnatural,gupta2021bert} or words of different types~\cite{sugawara2020assessing}, or to perform multi-hop reasoning~\cite{jiang-bansal-2019-avoiding}.
However, it is hard to rule out factors that may explain poor performance (e.g., distribution shift) or good performance (e.g., other ways to solve a problem).
Here, we examine an intrinsic property of the dataset (MDL) and provide a theoretical argument justifying why it is the right measure to use.

We use RDA to provide insights on a variety of datasets.
First, we verify that description length is reduced when we invoke a capability $f$ that is known to be helpful on a carefully-controlled synthetic task.
Next, we examine \hotpot~\cite{yang2018hotpotqa}, a benchmark for answering questions, where prior work has both claimed that decomposing questions into subquestions is helpful~\cite{min-etal-2019-multi,perez-etal-2020-unsupervised} and called such claims into question~\cite{min-etal-2019-compositional,jiang-bansal-2019-avoiding,chen-durrett-2019-understanding}.
RDA shows that subquestions are indeed helpful and exposes how evaluation procedures in prior work may have caused the value of question decomposition to be underestimated.
We then evaluate if explanations are useful for recognizing textual entailment using the e-SNLI dataset~\cite{camburu2018esnli}.
Both written explanations and decision-relevant keyword markings (``rationales'') are helpful, but rationales are more useful than explanations.
Lastly, we examine a variety of popular NLP tasks, evaluating the extent to which they require relying on word order, different types of words, and gender bias.
Overall, our results show that RDA can be used to answer a broad variety of questions about datasets.

\section{Rissanen Data Analysis}
\label{sec:Method}

How can we determine whether or not a certain capability $f(x)$ is helpful for building a good model of the data?
To answer this question, we view a dataset with inputs $x_{1:N}$ and labels $y_{1:N}$ as generated by a program that maps $x_n \rightarrow y_n$.
Let the length of the shortest such program $P$ be $\mathcal{L}(y_{1:N} | x_{1:N})$, the data's Kolmogorov complexity.
We view a capability as a function $f$ that maps $x_n$ to a possibly helpful output $f(x_n)$, with $\mathcal{L}(y_{1:N} | x_{1:N}, f)$ being the length of the shortest label-generating program when access to $f$ is given.
We say that $f$ is helpful exactly when:
\begin{equation}
\label{eqn:reducing-kc}
    \mathcal{L}(y_{1:N} | x_{1:N}, f) < \mathcal{L}(y_{1:N} | x_{1:N})
\end{equation}

\subsection{Minimum Description Length}

To use Eq.~\ref{eqn:reducing-kc} in practice, we need to find the shortest program $P$, which is uncomputable in general.
However, because $P$ is a program that generates $y_{1:N}$ given $x_{1:N}$, we can instead consider any compressed version of $y_{1:N}$, along with an accompanying decompression algorithm that produces $y_{1:N}$ given $x_{1:N}$ and the compressed $y_{1:N}$.
To find $\mathcal{L}$, then, we find the length of the maximally compressed $y_{1:N}$, or Minimum Description Length~\citep[MDL;][]{rissanen1978modeling}.
While MDL is not computable, just like Kolmogorov complexity, many methods have been proposed to estimate MDL by restricting the set of allowed compression algorithms~\citep[see][for an overview]{grunwald2004tutorial}.
These methods are all compatible with RDA, and here, we use online (or prequential) coding~\citep{rissanen1984universal,dawid1984present}, an effective method for estimating MDL when used with deep learning~\citep{blier2018description}.

\subsection{Online Coding}

To examine how much $y_{1:N}$ can be compressed, we look at the minimum number of bits (minimal codelength) needed by a sender Alice to transmit $y_{1:N}$ to a receiver Bob, when both share $x_{1:N}$.
Without loss of generality, we assume $y_n$ is an element from a finite set.
In online coding, Alice first sends Bob the learning algorithm $\mathcal{A}$, including the model architecture, trainable parameters $\theta$, optimization procedure, hyperparameter selection method, initialization scheme, random seed, and pseudo-random number generator.
Alice and Bob each initialize a model $p_{\theta_1}$ using the random seed and pseudo-random number generator, such that both models are identical.

Next, Alice sends each label $y_n$ one by one.
\citet{shannon1948mathematical} showed that there exists a minimum code to send $y_n$ with $-\log_2 p_{\theta_n}(y_n|x_n)$ bits when Alice and Bob share $p_{\theta_n}$ and $x_n$.
After Alice sends $y_n$, Alice and Bob use $\mathcal{A}$ to train a better model $p_{\theta_{n+1}}(y|x)$ on $(x_{1:n}, y_{1:n})$ to get shorter codes for future labels.
The codelength for $y_{1:N}$ is then:
\begin{equation}
\label{eqn:online-code}
    \mathcal{L}_p(y_{1:N} | x_{1:N}) = \sum_{n=1}^{N} -\log_2 p_{\theta_n}(y_n|x_n).
\end{equation}

Intuitively, $\mathcal{L}_p(y_{1:N} | x_{1:N})$ is the area under the ``online'' learning curve that shows how the cross-entropy loss goes down as training size increases.
Overall, Alice's message consists of $\mathcal{A}$ plus the label encoding ($\mathcal{L}_p (y_{1:N} | x_{1:N})$ bits).
When Alice and Bob share $f$, Alice's message consists of $\mathcal{A}$ plus $\mathcal{L}_p(y_{1:N} | x_{1:N}, f)$ bits to encode the labels with a model $p_{\theta}(y|x, f)$.
$f$ is helpful when the message is shorter with $f$ than without, i.e., when:
\begin{equation*}
    \mathcal{L}_p(y_{1:N} | x_{1:N}, f) < \mathcal{L}_p(y_{1:N} | x_{1:N})
\end{equation*}

\subsection{Implementation with Block-wise Coding}

The online code in Eq.~\ref{eqn:online-code} is expensive to compute.
It has a computational complexity that is quadratic in $N$ (assuming linear time learning), which is prohibitive for large $N$ and compute-intensive $\mathcal{A}$.
Following~\citet{blier2018description}, we upper bound online codelength by having Alice and Bob only train the model upon having sent $0 = t_0 < t_1 < \dots < t_S = N$ labels.
Alice thus sends all labels in a ``block'' $y_{t_s+1:t_{s+1}}$ at once using $p_{\theta_{t_s}}$, giving codelength:
\begin{equation*}
\label{eqn:block-wise-online-code-simple}
    \bar{\mathcal{L}}_p(y_{1:N} | x_{1:N}) = \sum_{s=0}^{S-1} \sum_{n=t_s+1}^{t_{s+1}} -\log_2 p_{\theta_{t_s}}(y_n|x_n)
\end{equation*}

Since $\theta_{t_0}$ has no training data, Alice sends Bob the first block using a uniform prior.

\paragraph{Alleviating the sensitivity to learning algorithm}

To limit the effect of the choice of learning algorithm $\mathcal{A}$, we may ensemble many model classes.
To do so, we have Alice train $M$ models of different classes and send the next block's labels using the model that gives the shortest codelength.
To tell Bob which model to use to decompress a block's labels, Alice also sends $\log_2 M$ bits per block $s = 1, \dots, S-1$, adding $(S-1)~\log_2 M$ to \desc.
In this way, \desc{} relies less on the behavior of a single model class.

\subsection{Experimental Setup}

To evaluate \desc, we first randomly sort examples in the dataset.
We use $S=9$ blocks where $t_0 = 0$ and $t_1 = 64 < \dots < t_S = N$ such that $\frac{t_{s+1}}{t_s}$ is constant (log-uniform spacing).
To train a model on the first $s$ blocks, we split the available examples into train (90\%) and dev (10\%) sets, choosing hyperparameters and early stopping epoch using dev loss (codelength).
We otherwise follow each model's training strategy and hyperparameter ranges as suggested by its original paper.
We then evaluate the codelength of the $(s+1)$-th block.
As a baseline, we show $\mathcal{H}(y)$, the codelength with the label prior $p(y)$ as $p_{\theta}$.

\desc{} is impacted by random factors such as the order of examples, model initialization, and randomness during training. Thus, we report the mean and std. error of \desc{} over 5 random seeds.
For efficiency, we only sweep over hyperparameters for the first random seed and reuse the best hyperparameters for the remaining seeds.
For all experiments, our code and reported codelengths are publicly available at \href{https://github.com/ethanjperez/rda}{https://github.com/ethanjperez/rda}, along with a short Python script to conduct RDA with your own models and datasets.

\section{Validating Rissanen Data Analysis}
\label{sec:Validating Rissanen Data Analysis}

Having described our setup, we now verify that \mbox{$\bar{\mathcal{L}}_p(y_{1:N} | x_{1:N}, f) < \bar{\mathcal{L}}_p(y_{1:N} | x_{1:N})$} holds in practice when using an $f$ that we know is helpful.
To this end, we use CLEVR~\cite{johnson2017clevr}, an synthetic, image-based question-answering (QA) dataset.
Many CLEVR questions were carefully designed to benefit from answering subquestions.
For example, to answer the CLEVR question ``\textit{Are there more cubes than spheres?}'' it helps to know the answer to the subquestions ``\textit{How many cubes are there?}'' and ``\textit{How many spheres are there?}''
We hypothesize that \desc{} decreases as we give a model answers to subquestions.

We test our hypothesis on three types of CLEVR questions.
``Integer Comparison'' questions ask to compare the numbers of two kinds of objects and have two subquestions (example above).
``Attribute Comparison'' questions ask to compare the properties of two objects (two subquestions).
``Same Property As'' questions ask whether or not one object has the same property as another (one subquestion).
Since CLEVR is synthetic, we obtain oracle answers to subquestions (``subanswers'') programmatically.
See Appendix \S\ref{ssec:CLEVR Details} for details.
We append subanswers to the question (in order) and evaluate MDL when providing 0-2 subanswers.

\paragraph{Model} 

We use the FiLM model from~\citet{perez2018film} which combines a convolutional network for the image with a GRU for the question~\citep[][]{cho-etal-2014-learning}.
The model minimizes cross-entropy loss (27-way classification).
We follow training strategy from~\citet{perez2018film} using the public code, except we train for at most 20 epochs (not 80), since we only train on subsets of CLEVR.

\paragraph{Results}

\begin{figure}[t]
	\centering
    \begin{subfigure}
        \centering
        \includegraphics[scale=.33]{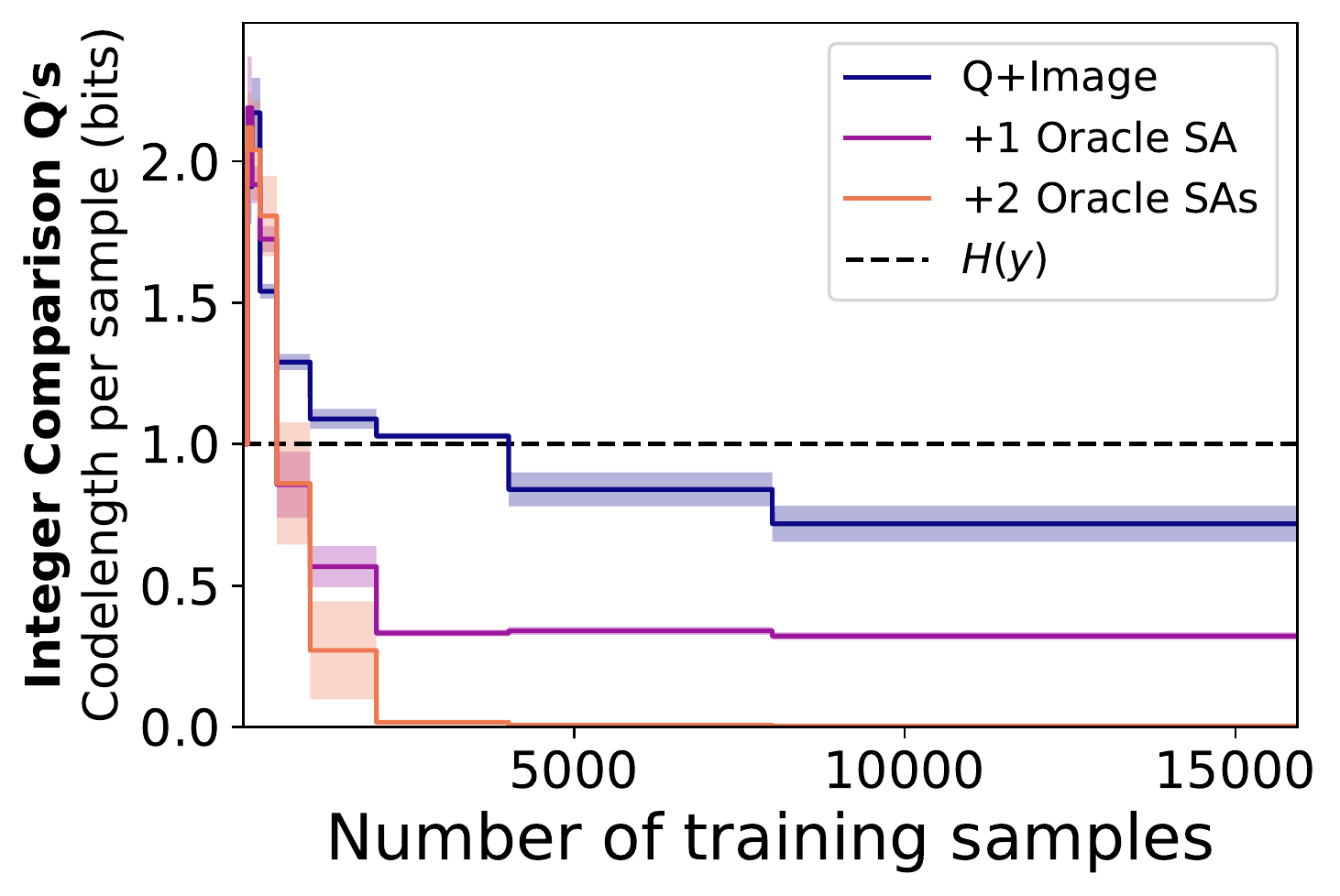}
    \end{subfigure}
    \begin{subfigure}
		\centering
        \includegraphics[scale=.29]{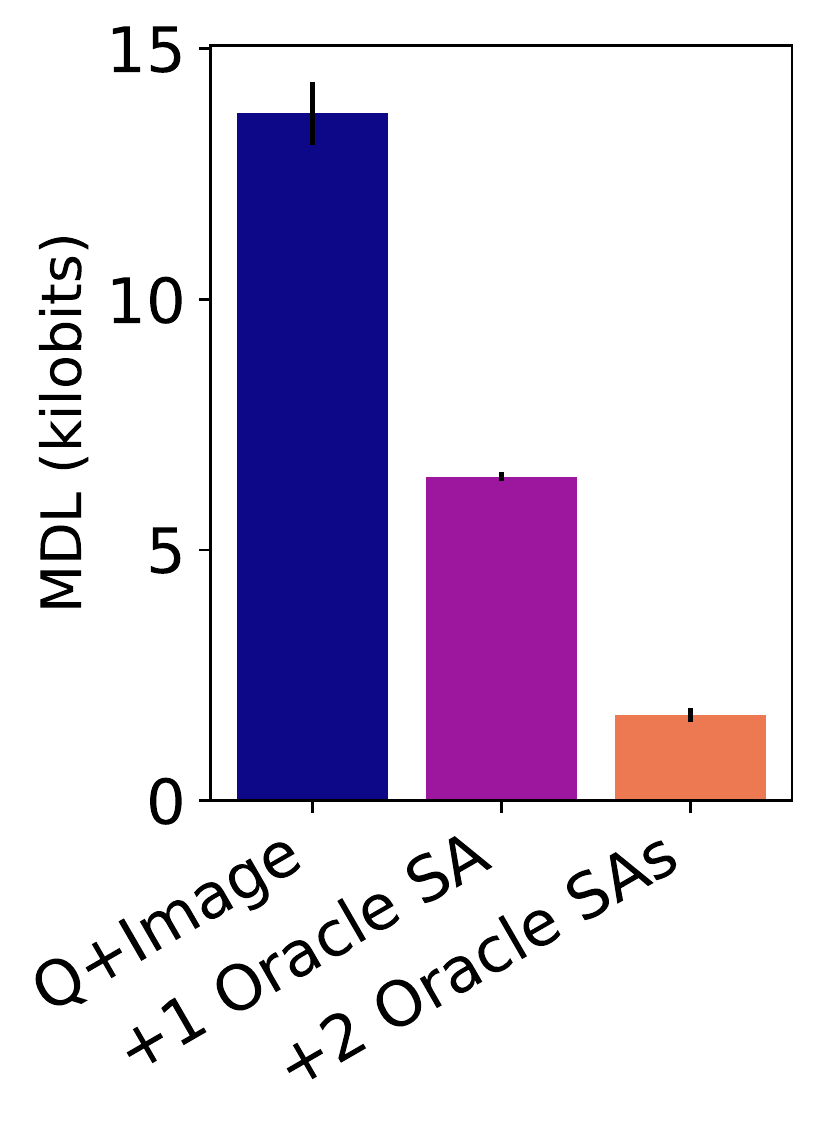}
    \end{subfigure}
    \\
    \begin{subfigure}
        \centering
        \includegraphics[scale=.33]{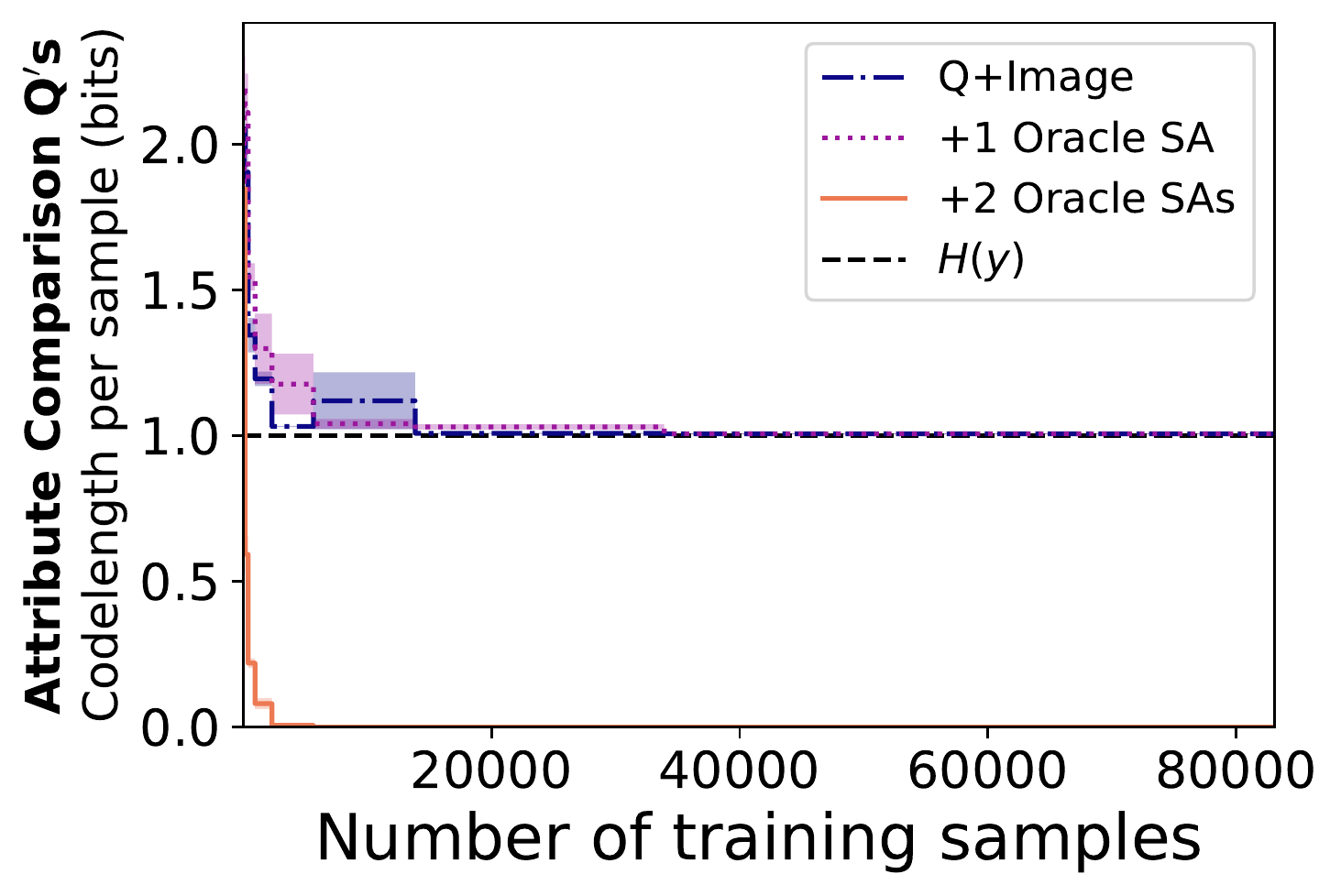}
    \end{subfigure}
    \begin{subfigure}
		\centering
        \includegraphics[scale=.29]{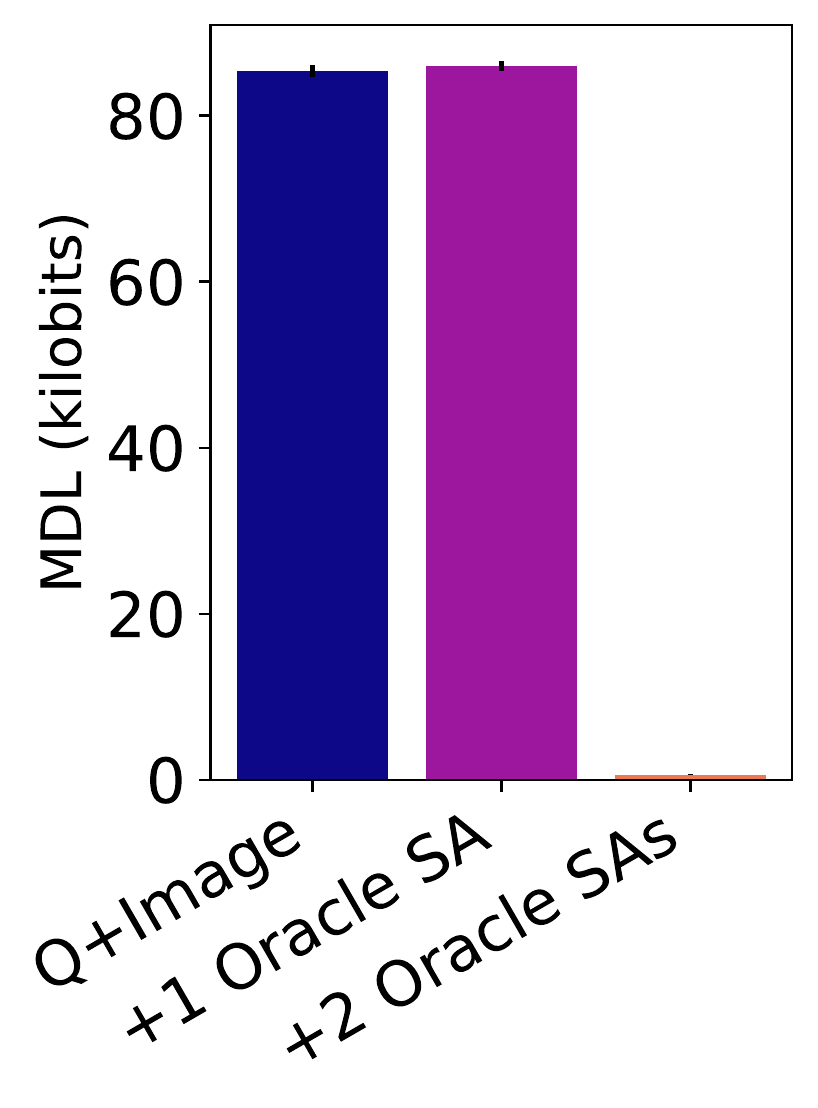}
    \end{subfigure}
    \\
    \begin{subfigure}
        \centering
        \includegraphics[scale=.32]{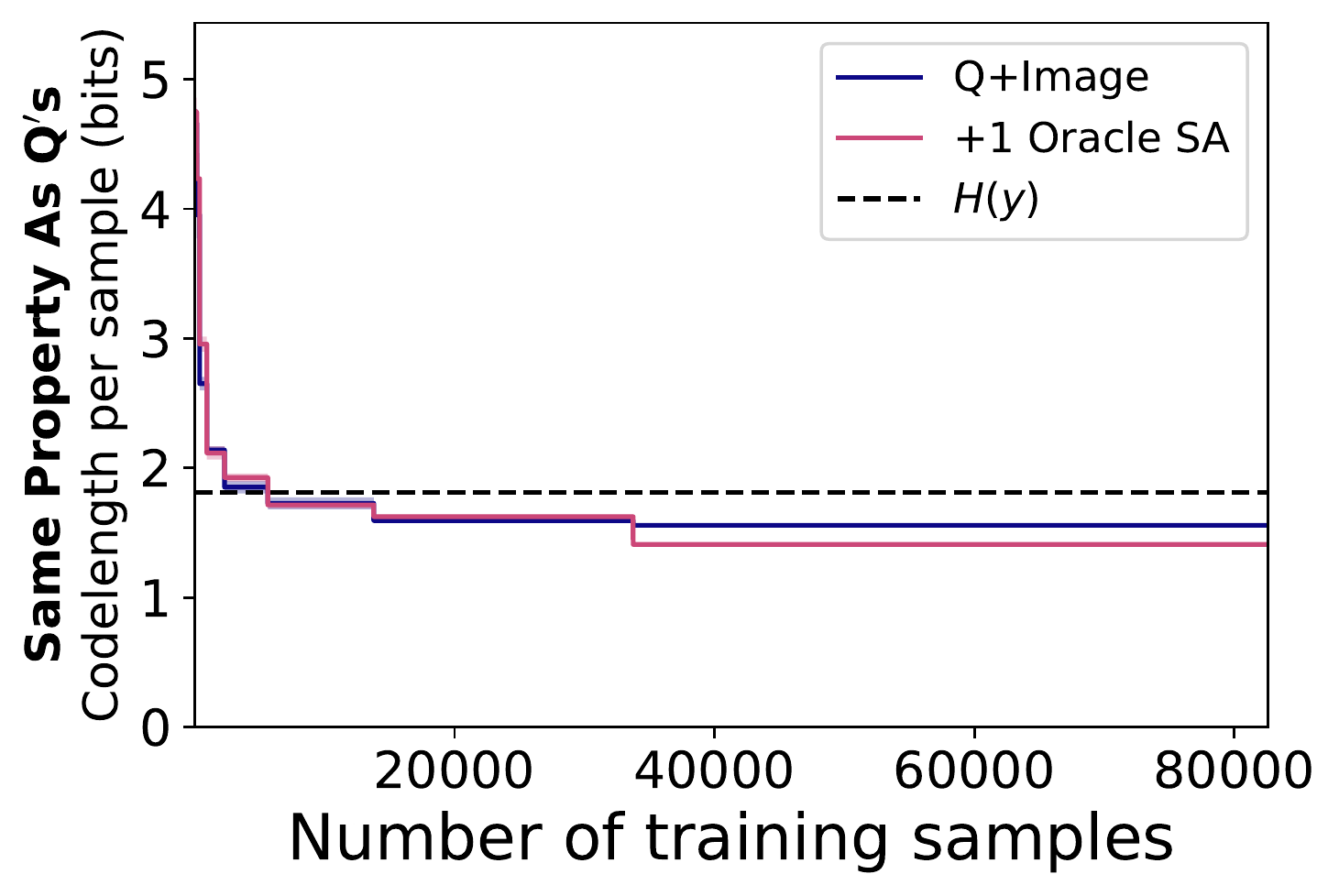}
    \end{subfigure}
    \begin{subfigure}
		\centering
        \includegraphics[scale=.29]{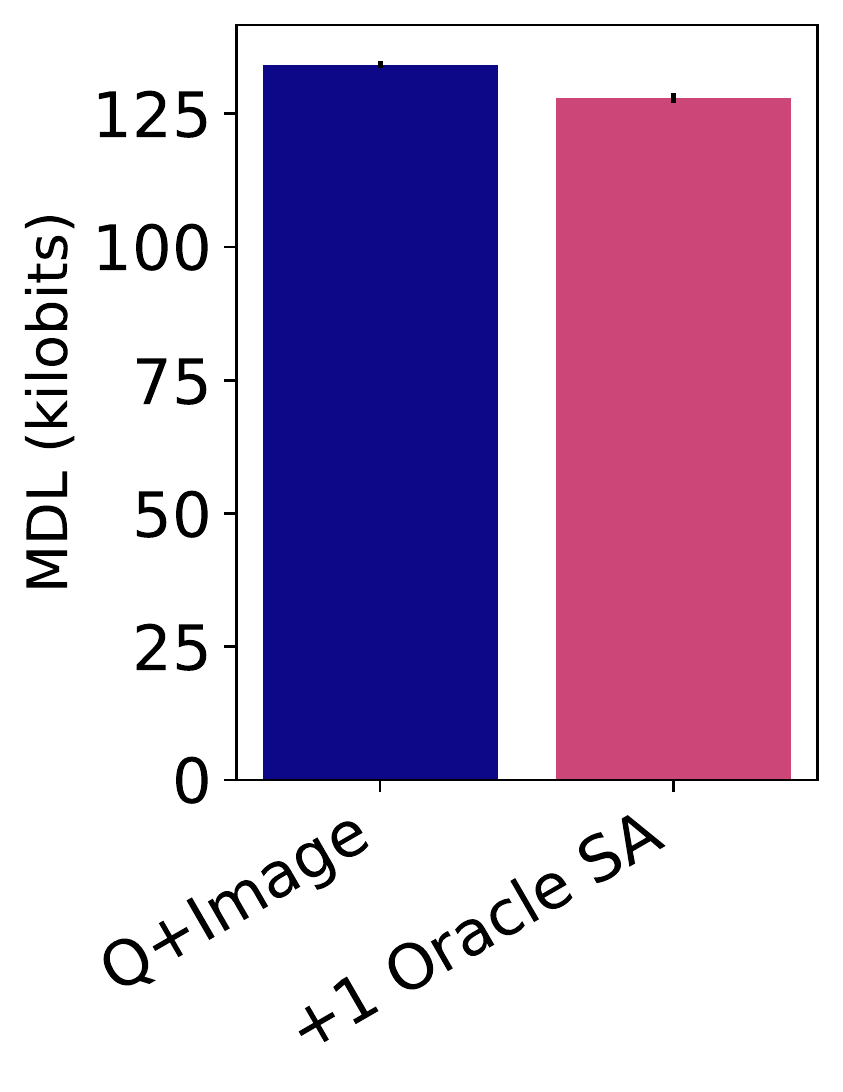}
    \end{subfigure}
    \caption{\textbf{Left}: Answer codelengths for different CLEVR question types with/without adding oracle answers to subquestions (``subanswers'') to the input. \textbf{Right}: Subanswers reduce \desc.}
    \label{fig:clevr}
\end{figure}

Fig.~\ref{fig:clevr} shows codelengths (left) and \desc{} (right).
For all question types, $\bar{\mathcal{L}}_p(y_{1:N} | x_{1:N}, f) < \bar{\mathcal{L}}_p(y_{1:N} | x_{1:N})$ when all oracle subanswers are given, as expected.
For ``Integer Comparison'' (top) and ``Attribute Comparison'' (middle), the reduction in \desc{} is larger than for ``Same Property As'' questions (bottom).
For comparison question types, the subanswers can be used without the image to determine the answer, explaining the larger decreases in \desc.
Our results align with our expectations about when answers to subquestions are helpful, empirically validating RDA.

\section{Examining Dataset Characteristics}
\label{sec:Examining Dataset Characteristics}
We now use RDA to answer pertinent open questions on various popular datasets.

\subsection{Is it helpful to answer subquestions?}
\label{ssec:hotpot}

\citet{yang2018hotpotqa} proposed \hotpot{} as a dataset that benefits from decomposing questions into subquestions, but recent work has called the benefit into doubt~\cite{min-etal-2019-compositional,jiang-bansal-2019-avoiding,chen-durrett-2019-understanding} while there is also evidence that decomposition helps~\cite{min-etal-2019-multi,perez-etal-2020-unsupervised}.
We use RDA to determine if subquestions and answers are useful.

\paragraph{Dataset}

\hotpot{} consists of crowdsourced questions (\textit{``Are Coldplay and Pierre Bouvier from the same country?''}) whose answers are intended to rely on information from two Wikipedia paragraphs.
The input consists of these two ``supporting'' paragraphs, 8 ``distractor'' paragraphs, and the question.
Answers are either \yes, \no, or a text span in an input paragraph.

\paragraph{Model} 

We use the \longformer~\cite{beltagy2020longformer}, a transformer~\cite{vaswani2017attention} modified to handle long inputs as in \hotpot.
We evaluate \desc{} for two models, the official \longformerbase{} initialized with pretrained weights trained on language modeling and another model with the same architecture that we train from scratch, which we refer to as \transformerbase.
We train the model to predict the span's start token and end token by minimizing the negative log-likelihood for each prediction.
We treat yes/no questions as span prediction as well by prepending \yes{} and \no{} to the input, following~\citet{perez-etal-2020-unsupervised}.
We use the implementation from~\citet{wolf-etal-2020-transformers}.
See Appendix \S\ref{ssec:Longformer} for hyperparameters.

\paragraph{Providing Subanswers}

We consider a subanswer to be a paragraph containing question-relevant information, because~\citet{perez-etal-2020-unsupervised} claimed that subquestions help by using a QA model to find relevant text.
We indicate up to two subanswers to the model by prepending ``$>$'' to the first subanswer paragraph and ``$\gg$'' to the second.

\paragraph{Choosing Subanswers}

We consider 5 methods for choosing subanswers.
First, we use the two supporting paragraphs as oracle subanswers.
Next, we consider the answers to subquestions from four different methods.
Three are unsupervised methods from~\citet{perez-etal-2020-unsupervised}: pseudo-decomposition (retrieval-based subquestions), seq2seq (subquestions from a sequence-to-sequence model), and ONUS (One-to-N Unsupervised Sequence transduction).
Last, we test the ability of a more recent, large language model~\citep[GPT3;][]{brown2020language} to generate subquestions using a few question-decomposition examples.
Since GPT3 is expensive to run, we use its generated subquestions as training data for a smaller T5 model~\cite{raffel2020exploring}, a ``Distilled Language Model'' (DLM, see Appendix \S\ref{ssec:Distilled Language Model Decompositions} for details).
To answer generated subquestions, we use the QA model from~\citet{perez-etal-2020-unsupervised}, an ensemble of two \robertalarge~\cite{liu2019roberta} models finetuned on SQuAD~\cite{rajpurkar-etal-2016-squad} to predict answer spans.
We use the paragraphs containing predicted answer spans to subquestions as subanswers.

\subsubsection{Results}

\begin{figure}[t]
	\centering
    \begin{subfigure}
        \centering
        \includegraphics[scale=.33]{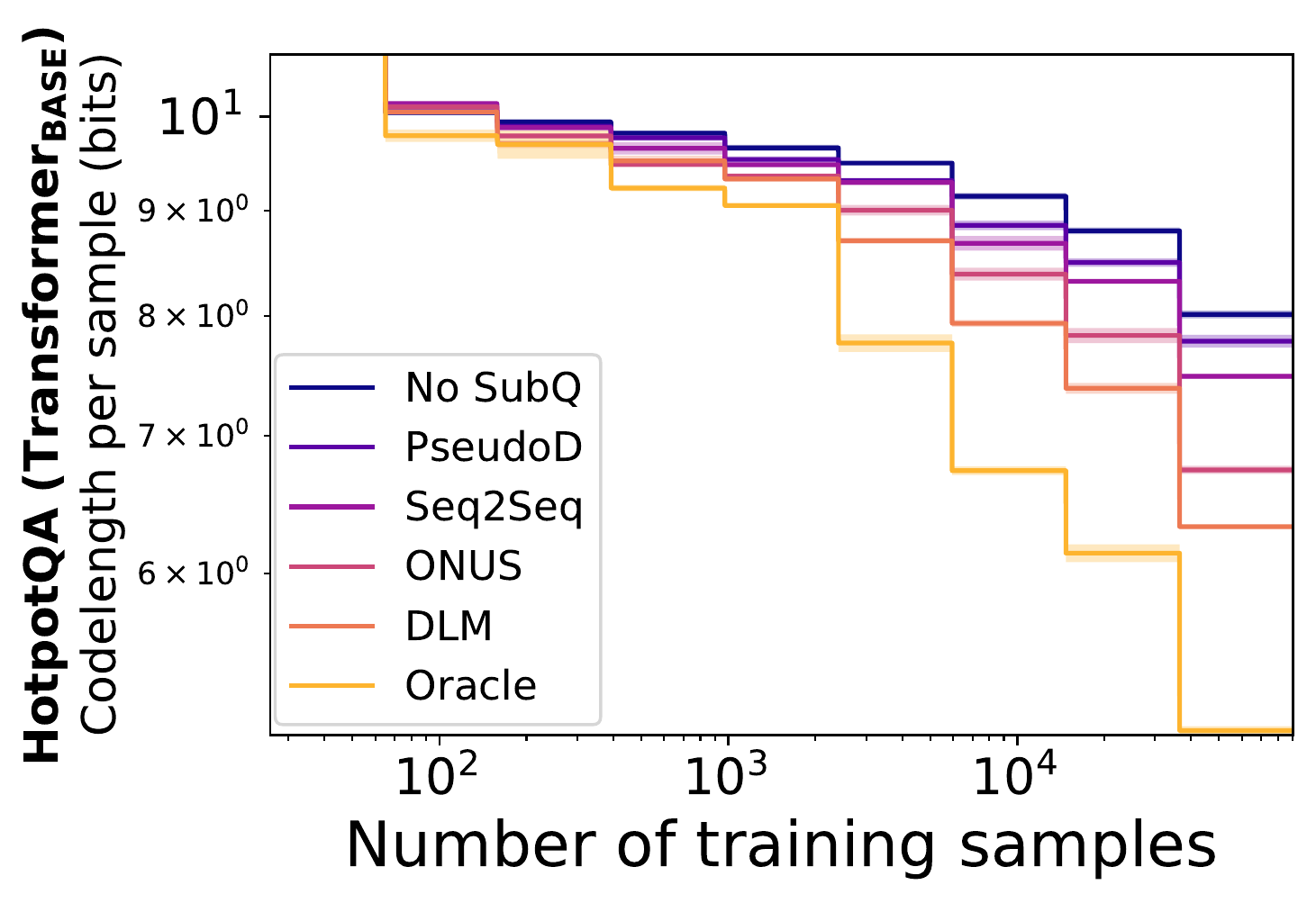}
    \end{subfigure}
    \begin{subfigure}
		\centering
        \includegraphics[scale=.28]{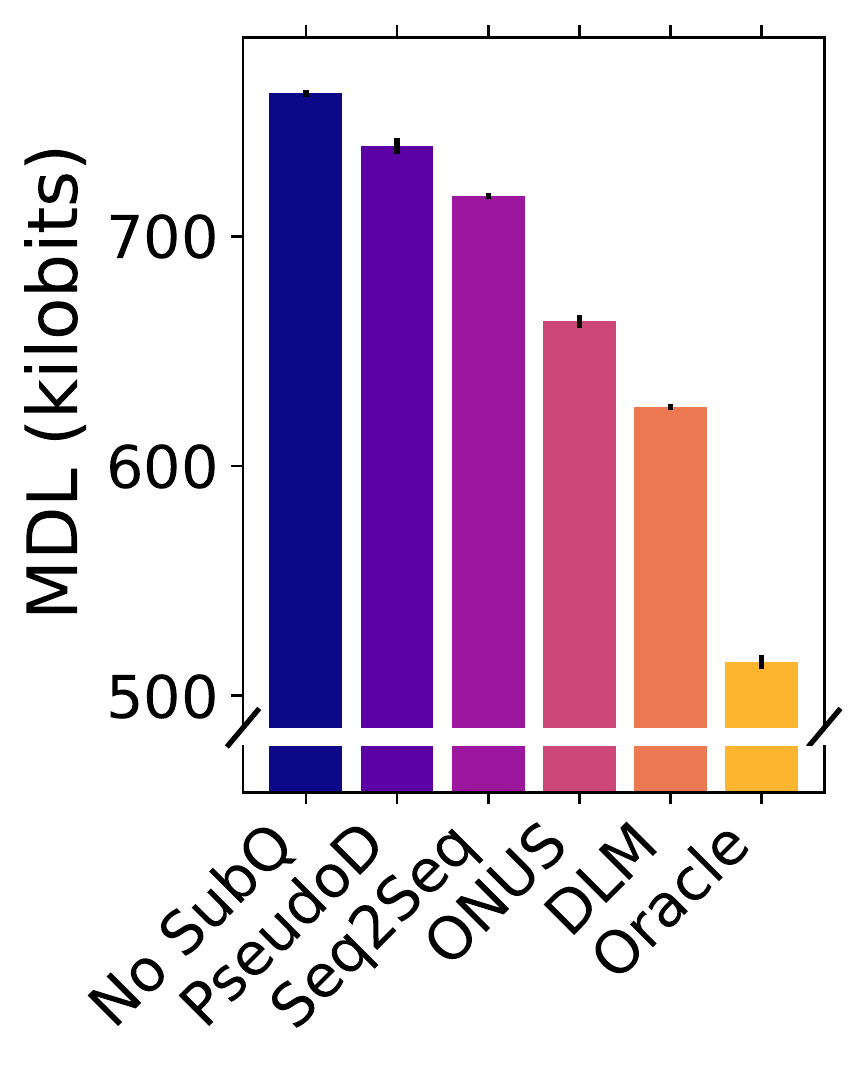}
    \end{subfigure}
    \\
    \begin{subfigure}
        \centering
        \includegraphics[scale=.33]{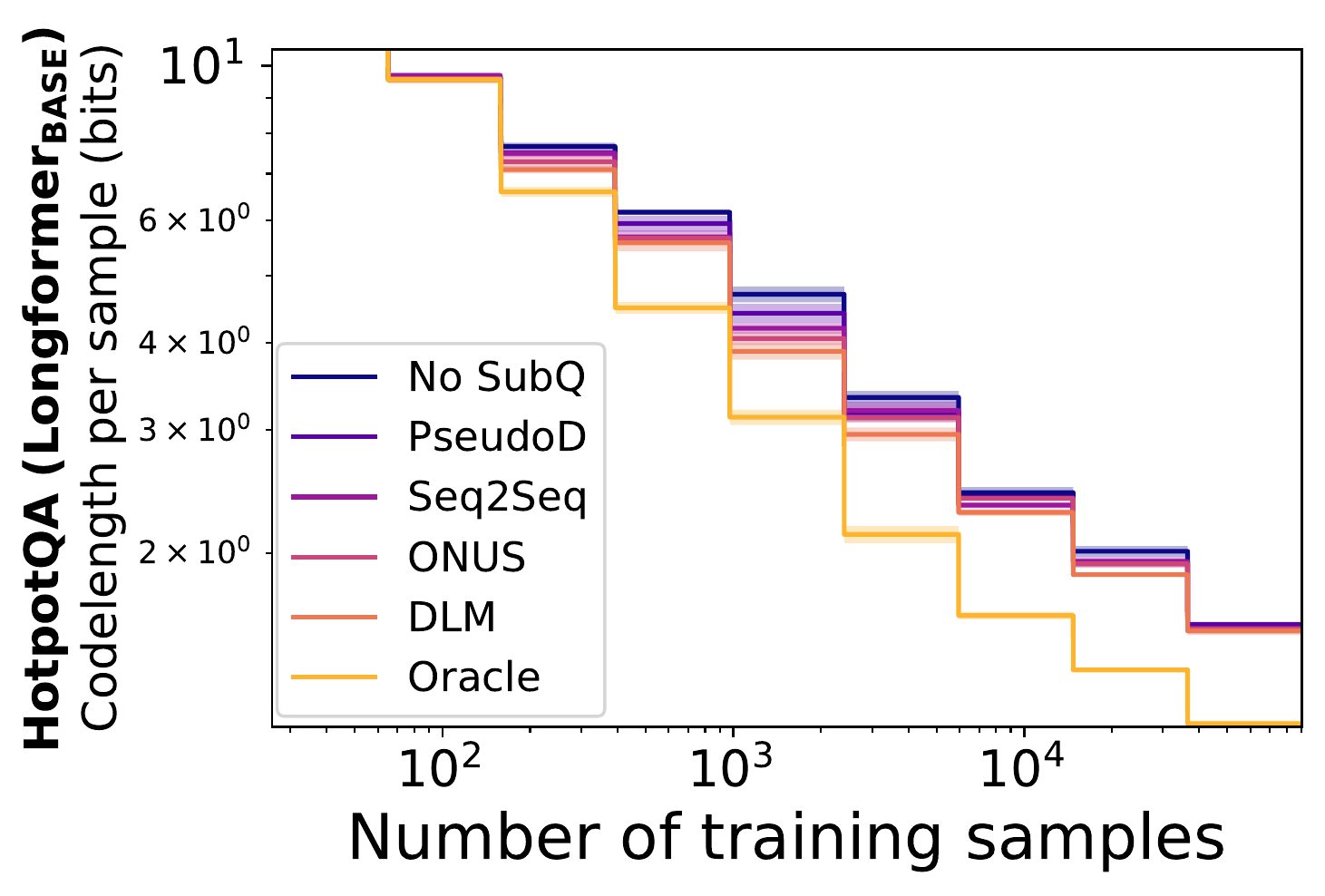}
    \end{subfigure}
    \begin{subfigure}
		\centering
        \includegraphics[scale=.28]{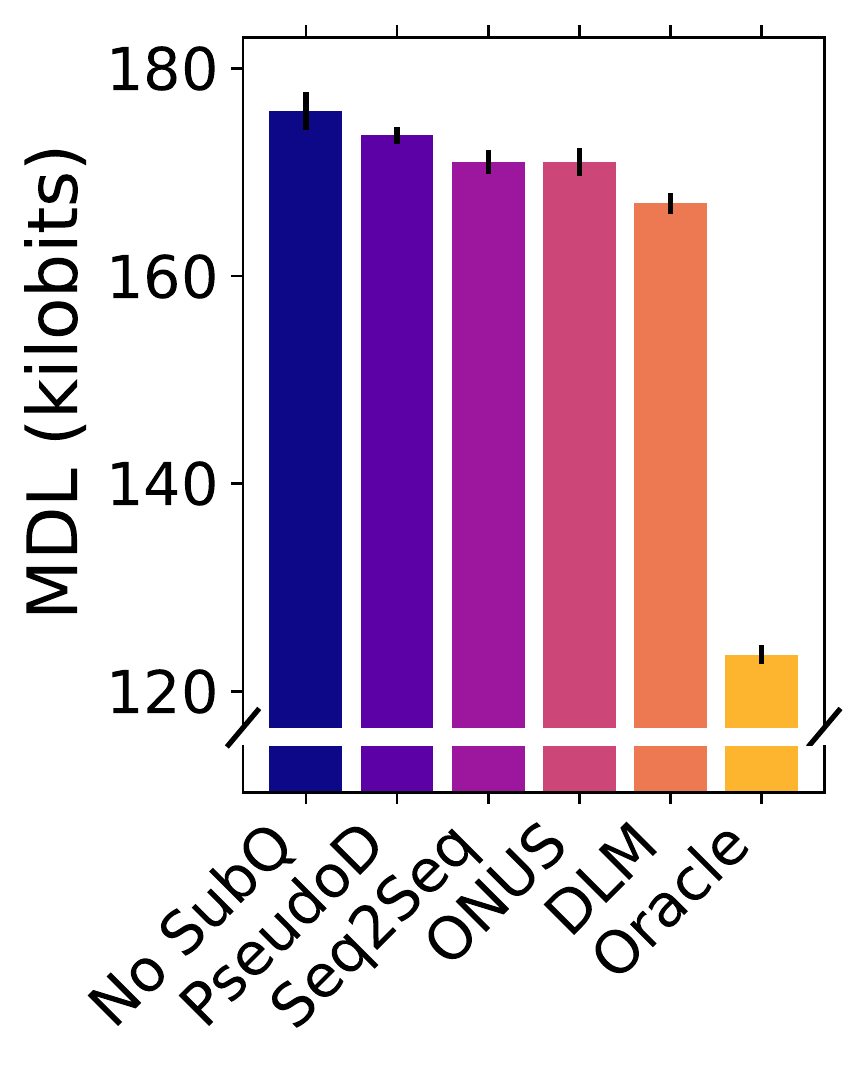}
    \end{subfigure}
    \caption{\textbf{Left}: Codelengths for \hotpot{} when encoding labels with \longformerbase{} trained from scratch (top) or pretrained weights (bottom), with the answers to subquestions (subanswers) from various decomposition methods. (Plots on log-log scale.) \textbf{Right}: \desc{} for decomposition methods when training from scratch (top) or pretrained weights (bottom). Subanswers help to compress the answers, especially when training from scratch, but with much room for improvement w.r.t. oracle subanswers.}
    \label{fig:hotpot}
\end{figure}

Fig.~\ref{fig:hotpot} shows codelengths (left) and \desc{} (right).
For \transformerbase{} (top), decompositions consistently and significantly reduces codelength and \desc{}.
Decomposition methods vary in how much they reduce \desc, ranked from worst to best as: no decomposition, Pseudo-Decomposition, Seq2Seq, ONUS, DLM, and oracle.
Overall, the capability to answer subquestions reduces program length, especially when subquestions and their answers are of high quality.

\begin{figure*}[t]
	\centering
    \begin{subfigure}
        \centering
        \includegraphics[scale=.43]{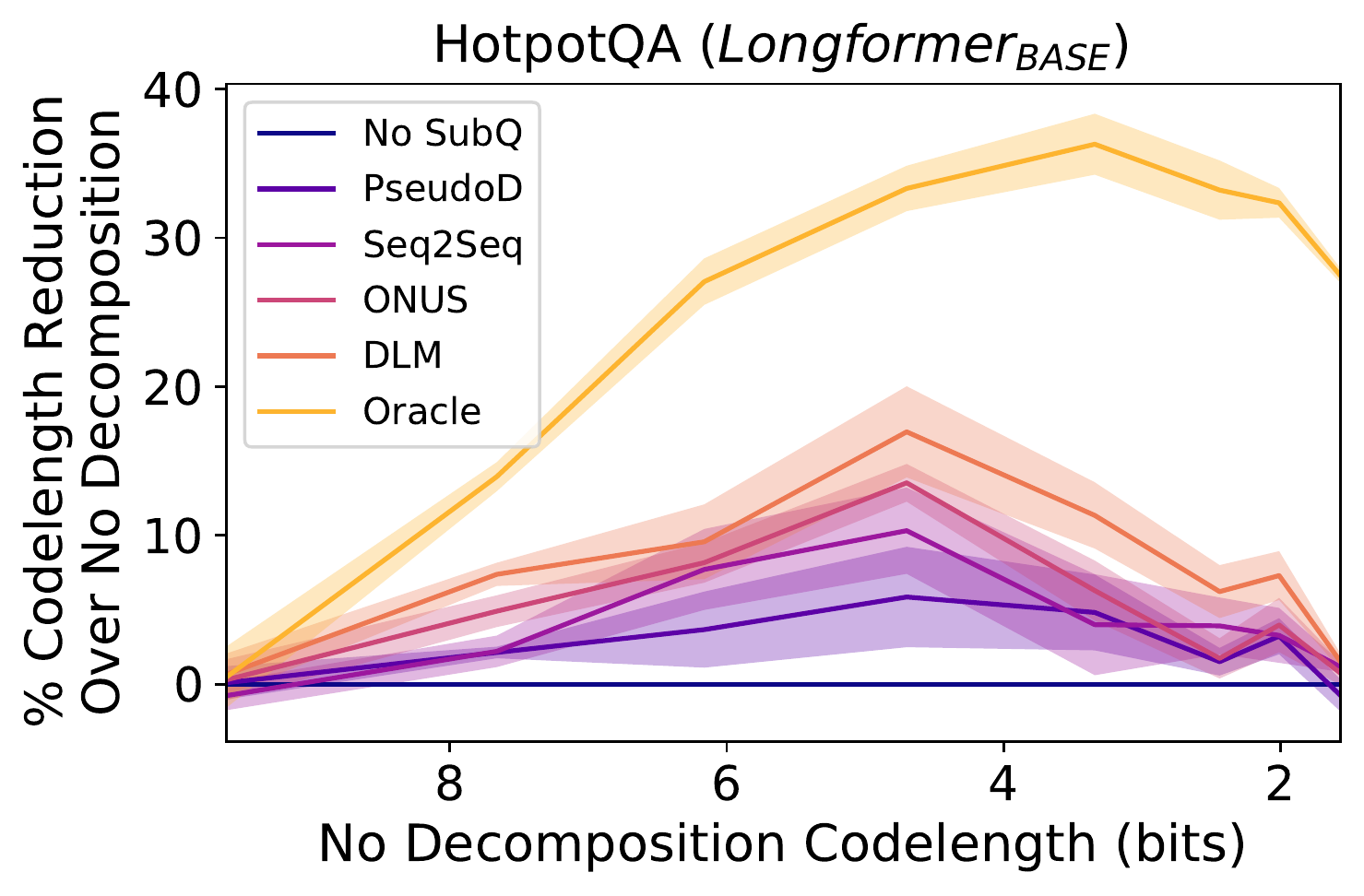}
    \end{subfigure}
    \begin{subfigure}
		\centering
        \includegraphics[scale=.41]{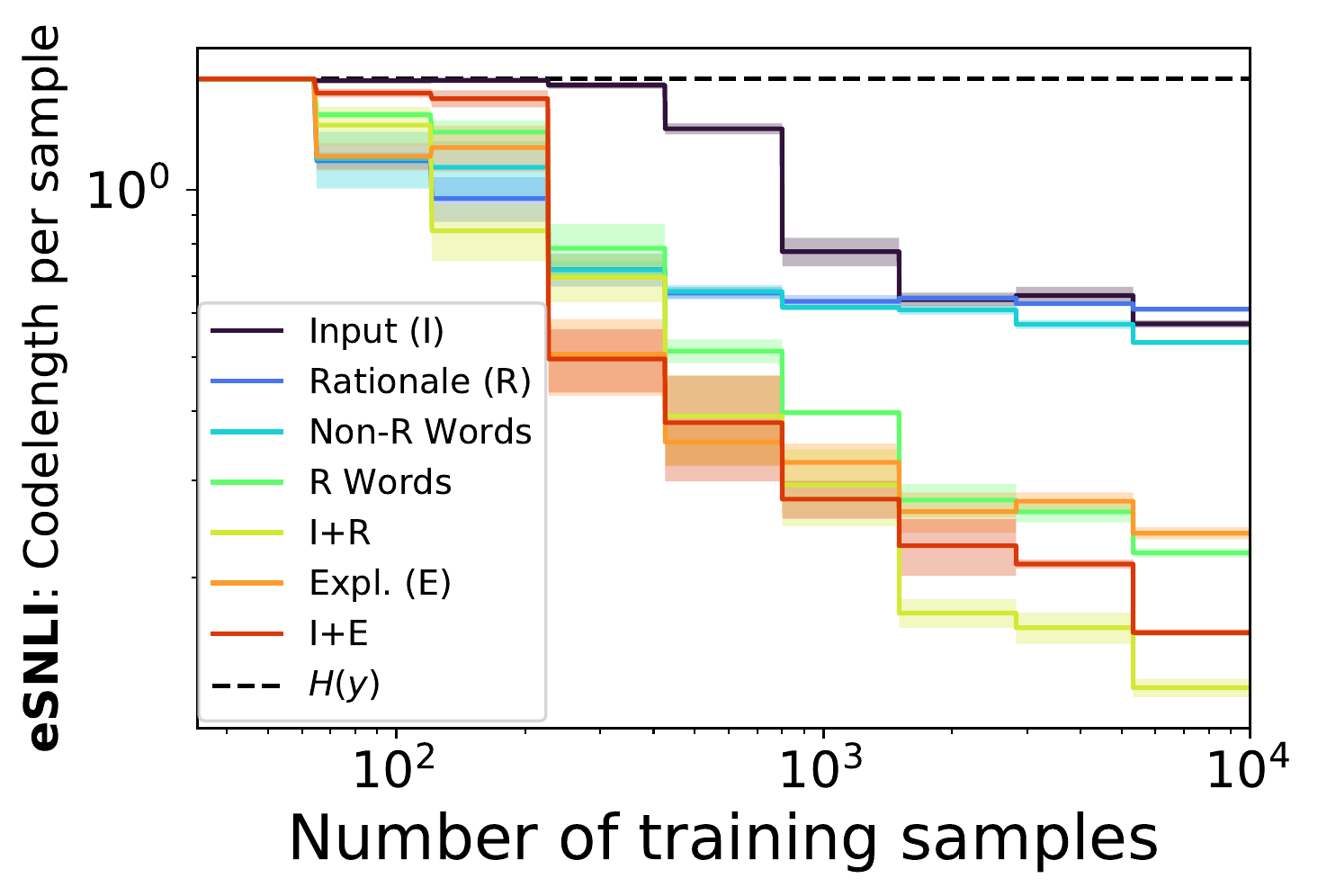}
    \end{subfigure}
    \begin{subfigure}
		\centering
        \includegraphics[scale=.33]{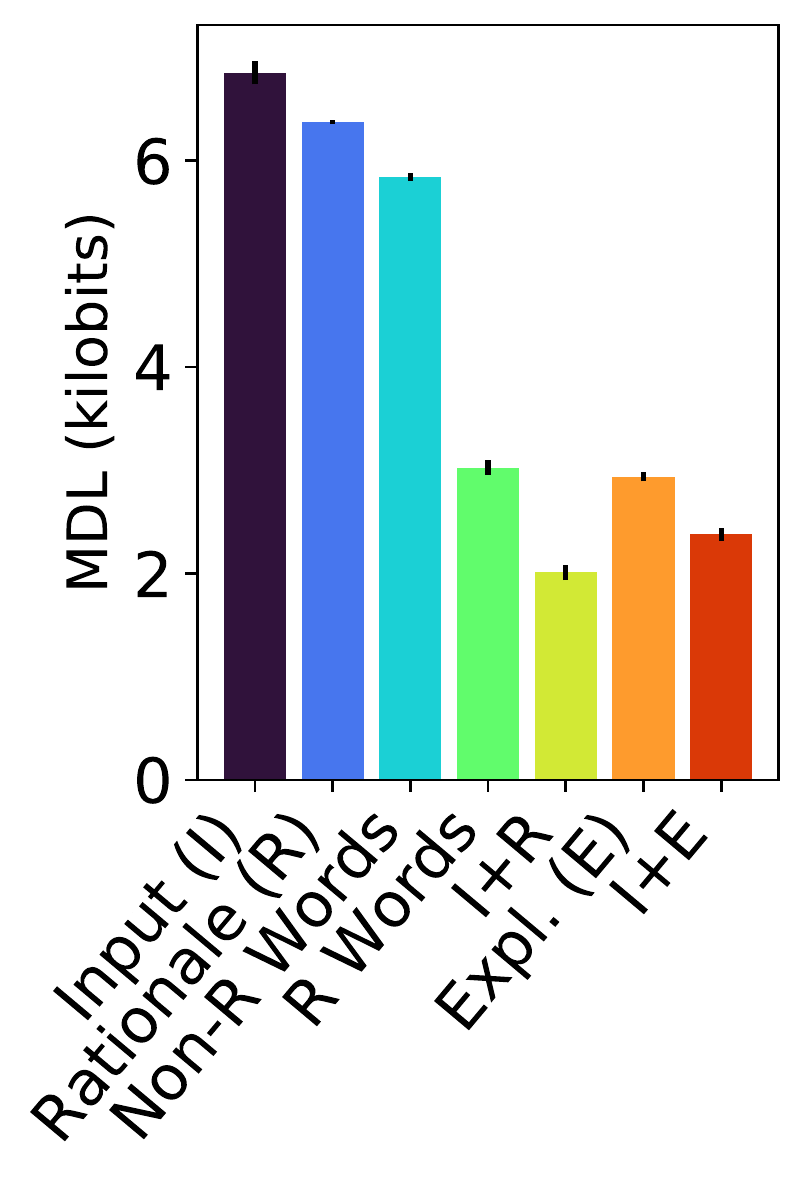}
    \end{subfigure}
    \caption{\textbf{Left}: On \hotpot, the reduction in codelength over the no-decomposition baseline from using subanswers from various decomposition methods (mean and std. err. for \longformerbase). \textbf{Middle}: Codelengths for e-SNLI with/without extractive rationales or written explanations. \textbf{Right}: On e-SNLI, \desc{} reduces significantly when rationales and explanations are given alongside the input.}
    \label{fig:hotpot-esnli}
\end{figure*}

For \longformerbase{} (Fig.~\ref{fig:hotpot} bottom), all decomposition methods also reduce codelength and \desc, though to a lesser extent.
To examine why, we plot the codelength reduction from decomposition against the original codelength for \longformerbase{} in Fig.~\ref{fig:hotpot-esnli} (left).
As the original codelength decreases, the benefit from decomposition increases, until the no-decomposition baseline reaches a certain loss, at which point the benefit from decomposition decreases.
We hypothesize that a certain, minimum amount of task understanding is necessary before decompositions are useful (see Appendix \S\ref{ssec:Is it helpful to answer subquestions? Details} for similar findings with \transformerbase).
However, as loss decreases, the task-relevant capabilities can be learned from the data directly, without decomposition.

Our finding suggests that decompositions help disproportionately in the high- or mid- loss regimes rather than the low-loss regime, where QA systems are usually evaluated (i.e., when training on all examples).
The limited value in low-loss regimes occurs because models approach the same, minimum loss $H(y|x)$ in the limit of dataset size.
Our observation partly explains why a few earlier studies~\citep{min-etal-2019-compositional,chen-durrett-2019-understanding}, which only evaluated final performance, drew the conclusion that \hotpot{} does not benefit much from multi-step reasoning or question decomposition.
In contrast, \desc{} actually does capture differences in performance across data regimes, demonstrating that RDA is the right approach going forward, especially given the growing interest in few-shot data regimes~\cite{lake2017building,brown2020language}

\subsection{Are Explanations and Rationales Useful?}
\label{ssec:esnli}

Recent work has proposed methods that give reasons for an answer before predicting the answer to improve accuracy.
Such reasons include written explanations~\cite{camburu2018esnli,rajani-etal-2019-explain,wiegreffe2020measuring} or locating task-relevant input words~\cite{zhang-etal-2016-rationale,perez-etal-2019-finding}.
As a testbed, these studies often use natural language inference (NLI) -- checking if a premise entails or contradicts (or neither) a hypothesis.
To explore if this direction is promising, we evaluate whether providing a reason is a useful capability, using NLI as a case study.

\paragraph{Dataset} 

We use the e-SNLI~\cite{camburu2018esnli} dataset, which annotated each example in SNLI~\cite{bowman-etal-2015-large} with two forms of reasons: an extractive rationale that marks entailment-relevant words and a written explanation of the right answer. We randomly sample 10k examples from e-SNLI to examine the usefulness of rationales and explanations.
To illustrate, e-SNLI contains an example of contradiction where the premise is ``\textit{A man and a woman are dancing in the \textbf{crowd}.}'' and the hypothesis is ``\textit{A man and woman dance \textbf{alone}.}''
The rationale is bolded, and the explanation is ``\textit{Being in a crowd means not alone.}''

\paragraph{Adding explanations and rationales}
We view rationales and explanations as generated by a function $f$ on the input.
To test if $f$ reduces \desc, we add the rationale by surrounding each entailment-relevant word with asterisks, and we add the explanation before the hypothesis, separated by a special token.
For comparison, we evaluate \desc{} when including only the explanation as input and only the rationale patterns as input.
For the latter, we use the rationale without the actual premise and hypothesis words by replacing each rationale word with ``*'' and other words with ``\textunderscore''.

\paragraph{Model} We use an ensemble model composed of the following model classes: FastText Bag-of-Words~\cite{joulin2017bag}, transformers~\cite{vaswani2017attention} trained from scratch (110M and 340M parameter versions), \bartbase{}~\citep[encoder-decoder;][]{lewis-etal-2020-bart}, \albertbase{}~\citep[encoder-only;][]{lan2020albert}, \robertabase{} and \robertalarge{}~\citep[encoder-only;][]{liu2019roberta} and the distilled version \distilroberta~\cite{sanh2019distilbert}, and \gpt~\citep[decoder-only;][]{radford2019language} and \distilgpt~\cite{sanh2019distilbert}.
For each model, we minimize cross-entropy loss and tune softmax temperature\footnote{Search over $[10^{-1}, 10^{2}]$, 1k log-uniformly spaced trials.} on dev to alleviate overconfidence on unseen examples~\cite{guo2017calibration,desai-durrett-2020-calibration}.
We follow each models' official training strategy and hyperparameter sweeps (Appendix \S\ref{ssec:Ensemble Model}), using the official codebase for FastText\footnote{\href{https://github.com/facebookresearch/fastText}{https://github.com/facebookresearch/fastText}} and HuggingFace Transformers~\cite{wolf-etal-2020-transformers} with PyTorch Lightning~\cite{falcon2019pytorch} for other models.

\subsubsection{Results}

Fig.~\ref{fig:hotpot-esnli} shows codelengths (middle) and \desc{} (right).
Adding rationales to the input greatly reduces \desc{} compared to using the normal input (``Input (I)'') or rationale markings without input words (``Rationale (R)''), suggesting that rationales complement the input.
The reduction comes from focusing on rationale words specifically.
We see almost as large \desc{} reductions when only including rationale-marked words and masking non-rationale words (``R Words'' vs. ``I+R'').
In contrast, we see little improvement over rationale markings alone when using only non-rationale words with rationale words masked (``Rationale (R)'' vs.``Non-R Words'').
Our results show that for NLI, it is useful to first determine task-relevant words, suggesting future directions along the lines of~\citet{zhang-etal-2016-rationale,perez-etal-2019-finding}.

Similarly, explanations greatly reduce \desc~(Fig.~\ref{fig:hotpot-esnli} right, rightmost two bars), especially when the input is also provided.
This finding shows that explanations, like rationales, are also complementary to the input.
Interestingly, adding rationales to the input reduces \desc{} more than adding explanations, suggesting that while explanations are useful, they are harder to use for label compression than rationales.

\subsection{Examining Text Datasets}

So far, we used RDA to determine when adding input features helps reduce label description lengths.
Similarly, we evaluate when removing certain features increases description length, to determine what features help achieve a small \desc.
Here, we view the ``original'' input as having certain features missing, and we evaluate the utility of a capability $f$ that recovers the missing features to return the normal task input.
If $f$ reduces the label-generating program length, then it is useful to have access to $f$ (the ablated features).
To illustrate, we evaluate the usefulness of different kinds of words and of word order on the General Language Understanding Evaluation benchmark~\citep[GLUE;][]{wang2018glue}, a central evaluation suite in NLP, as well as SNLI and Adversarial NLI~\citep[ANLI;][]{nie-etal-2020-adversarial}.


\paragraph{Datasets}
GLUE consists of 9 tasks (8 classification, 1 regression).\footnote{See Appendix \S\ref{ssec:GLUE Details} for details on GLUE and Appendix \S\ref{ssec:Regression} for details on regression.}
CoLA and SST-2 are single-sentence classification tasks.
MRPC, QQP, and STS-B involve determining if two sentences are similar or paraphrases of each other.
QNLI, RTE, MNLI, and WNLI are NLI tasks (we omit WNLI due to its size, 634 training examples).
ANLI consists of NLI data collected in three rounds, where annotators wrote hypotheses that fooled state-of-the-art NLI models trained on data from the previous round.
We consider each round as a separate dataset, to examine how NLI datasets have evolved over time, from SNLI to MNLI to ANLI$_1$, ANLI$_2$, and ANLI$_3$.

\paragraph{Experimental Setup}
Following our setup for e-SNLI (\S\ref{ssec:esnli}), we use the 10-model ensemble and evaluating \desc{} on up to 10k examples per task.

\subsubsection{The usefulness of part-of-speech words}
\label{ssec:pos}

\begin{figure}[t]
	\centering
    \includegraphics[scale=0.52]{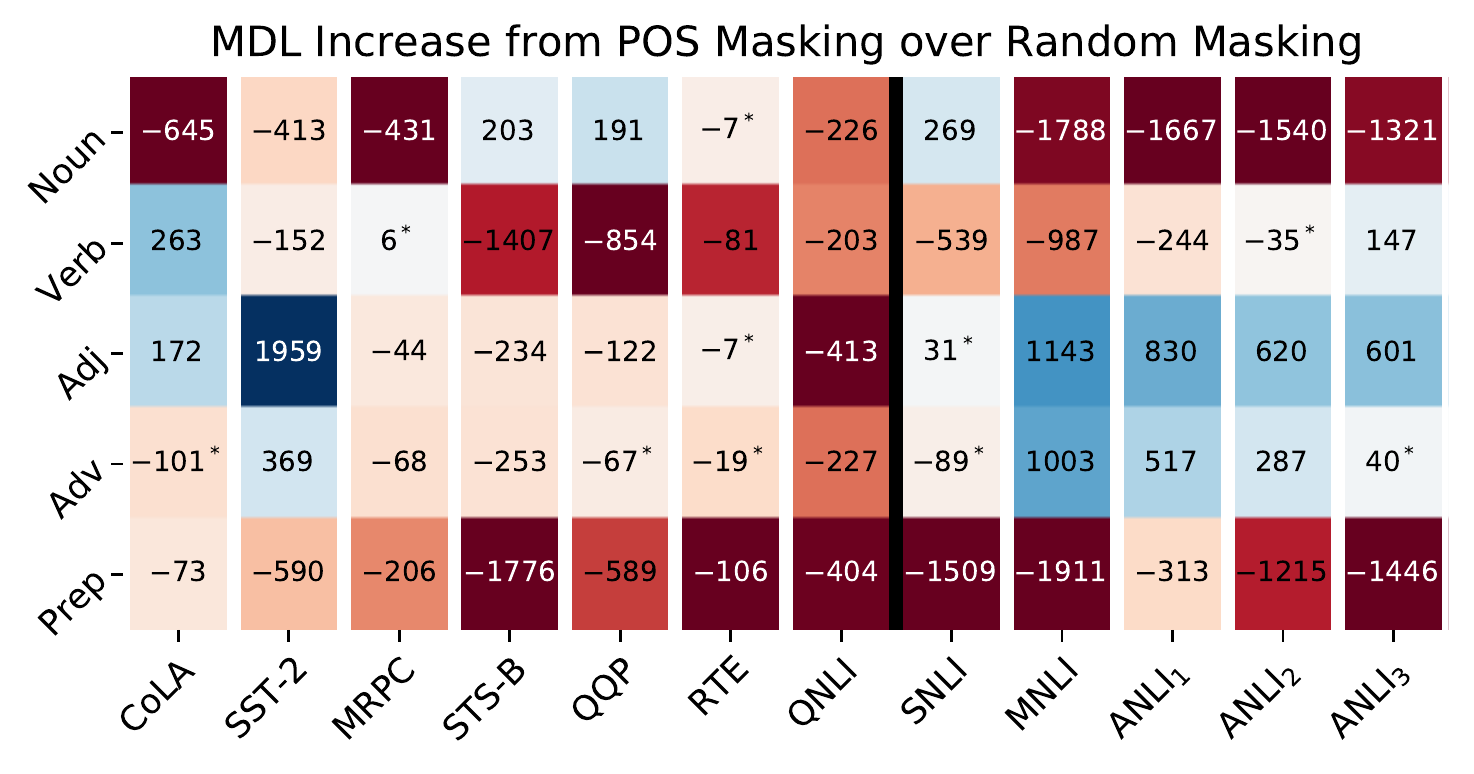}
    \caption{The importance of different POS words, given by $\desc_{-\text{POS}} - \desc_{-\text{Random}}$. 0 indicates that words of a given POS are as important as randomly-chosen words, while $>0$ and $<0$ indicate greater and lesser importance than randomly-chosen words, respectively. (*) indicates within std. error of 0. Color is normalized by column (dataset).}
    \label{fig:glue-pos-ablations}
\end{figure}

We consider the original input to be the full input with words of a certain POS masked out (with ``\_'') and evaluate the utility of a function $f$ that fills in the masked words.
To control for the number of words masked, we restrict $f$ such that it returns a version of the input with the same proportion of words masked, chosen uniformly at random.
If $f$ is useful, then words of a given type are more useful for compression than randomly-chosen input words.
In particular, we report the difference between \desc{} when (1) words of a given POS are masked and (2) the same fraction of words are masked uniformly at random: $\desc_{-\text{POS}} - \desc_{-\text{Random}}$.
We evaluate nouns, verbs, adjectives, adverbs, and prepositions.\footnote{We use POS tags from spaCy's large English model~\cite{honnibal2017spacy}. We omit other POS, as they occur less frequently and masking them did not greatly impact \desc{} in preliminary experiments.}

We show results in Figure~\ref{fig:glue-pos-ablations}.
Adjectives are much more useful than other POS for SST-2, a sentiment analysis task where relevant terms are evidently descriptive words (e.g., ``the service was \textit{terrible}'').
For CoLA, verbs play an important role in determining if a sentence is linguistically acceptable, likely due to the many examples evaluating verb argument structure (e.g., ``The toast burned.'' vs. ``The toast buttered.'').
Other tasks (MRPC, RTE, and QNLI) do not rely significantly on any one POS, suggesting that they require reasoning over multiple POS in tandem.
Nouns are consistently less useful on NLI tasks, suggesting that NLI datasets should be supplemented with knowledge-intensive tasks like open-domain QA that rely on names and entities, in order to holistically evaluate language understanding.
Prepositions are not important for any GLUE task, suggesting where GLUE can be complemented with other tasks and illustrating how RDA can be used to help form comprehensive benchmarks in the future.

\subsubsection{How useful are other word types?}

\citet{sugawara2020assessing} hypothesized other word types that may be useful for NLP tasks.
We use RDA to assess their usefulness as we did above (see Appendix \S\ref{sec:Additional Experiments} for details).
GLUE tasks vary in their reliance on ``content'' words.
Logical words like \textit{not} and \textit{every} are particularly important for MNLI which involves detecting logical entailment.
On the other hand, causal words (e.g., \textit{because}, \textit{since}, and \textit{therefore}) are not particularly useful for GLUE.

\subsubsection{Do Datasets Suffer from Gender Bias?}
\label{ssec:gender-bias}

\begin{figure}[t!]
	\centering
    \includegraphics[scale=0.497]{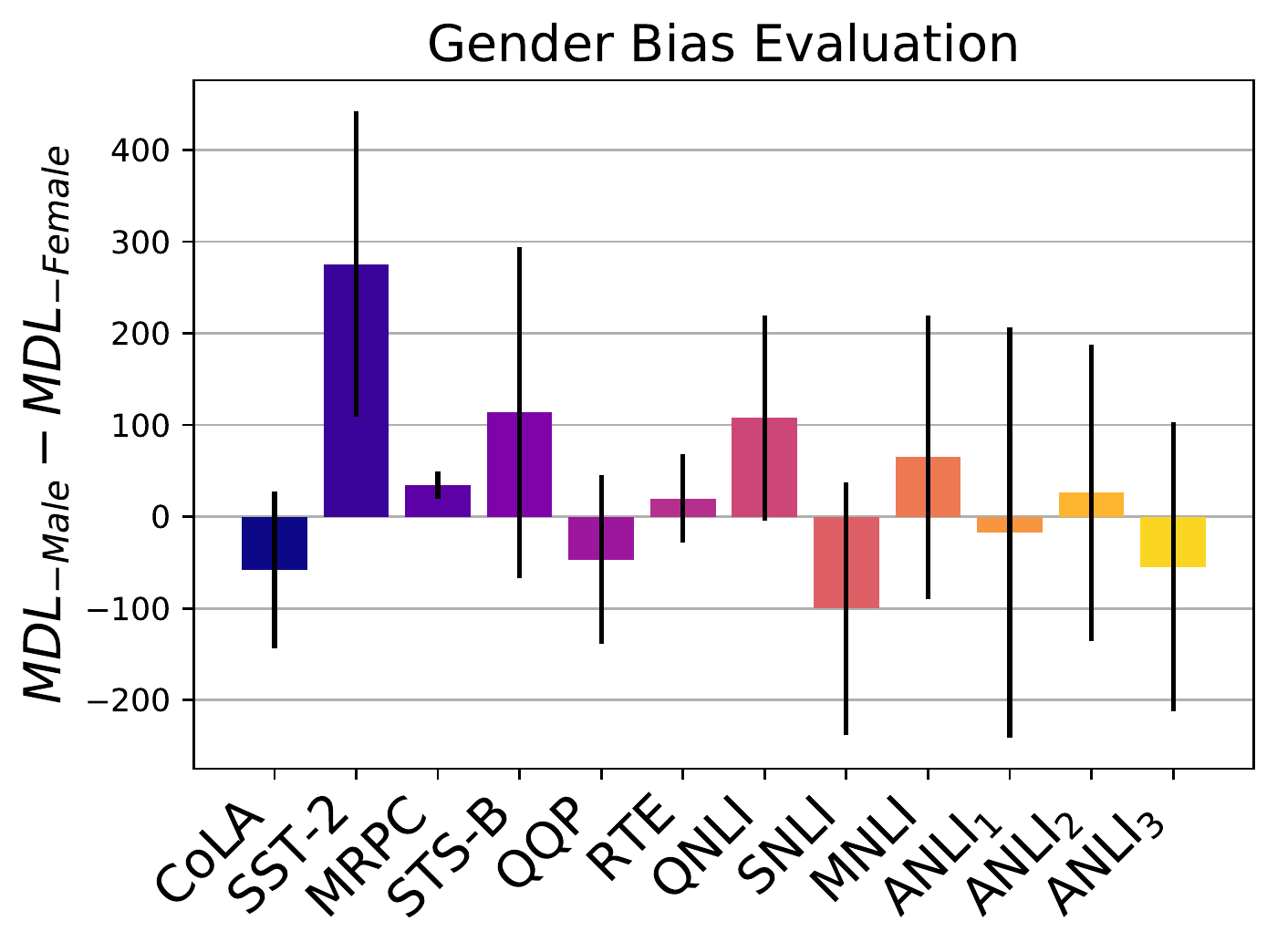}
    \caption{\textbf{Gender Bias}: \desc{} when masking masculine vs. feminine words (mean and std. err. over 5 random seeds). Values above zero (vs. below zero) indicate that male-gendered words (vs. female-gendered words) are more important for compressing labels. SST-2 shows the largest bias (male-favored).}
    \label{fig:gender}
\end{figure}

Gender bias in data is a prevalent issue in machine learning~\cite{bolukbasi2016man,blodgett-etal-2020-language}.
For example, prior work found that machine learning systems are worse at classifying images of women~\cite{phillips2000feret,buolamwini2018gender}, at speech recognition for women and speakers from Scotland~\cite{tatman-2017-gender}, and at POS tagging for African American vernacular~\cite{jorgensen-etal-2015-challenges}.
RDA can be used to diagnose such biases.
Here, we do so by masking male-gendered words and evaluating the utility of an oracle function $f$ that reveals male-gendered words while masking female-gendered words.
If $f$ is useful and $\desc_{-\text{Male}}-\desc_{-\text{Female}}>0$, then masculine words are more useful than feminine words for the dataset (gender bias).
We use male and female word lists from~\citet{dinan-etal-2020-queens,dinan-etal-2020-multi}.
The two lists are similar in size ($\sim$530 words each) and POS distribution (52\% nouns, 29\% verbs, 18\% adjectives), and the male- and female- gendered words occur with similar frequency.
See Appendix \S\ref{sec:Additional Experiments} for experiments controlling for word frequency.

Fig.~\ref{fig:gender} shows the results. Masculine words are more useful for SST-2 and MRPC while no GLUE datasets have feminine words as more useful.
For SST-2, feminine words occur more frequently than masculine words (2.7\% vs. 2.2\%, evenly distributed across class labels), suggesting that RDA uncovers a gender bias that word counts do not.
This result highlights the practical value of RDA in uncovering where evaluation benchmarks under-evaluate the performance of NLP systems on text related to different demographic groups.

\subsubsection{How useful is word order?}

\begin{figure}[t!]
	\centering
    \includegraphics[scale=0.45]{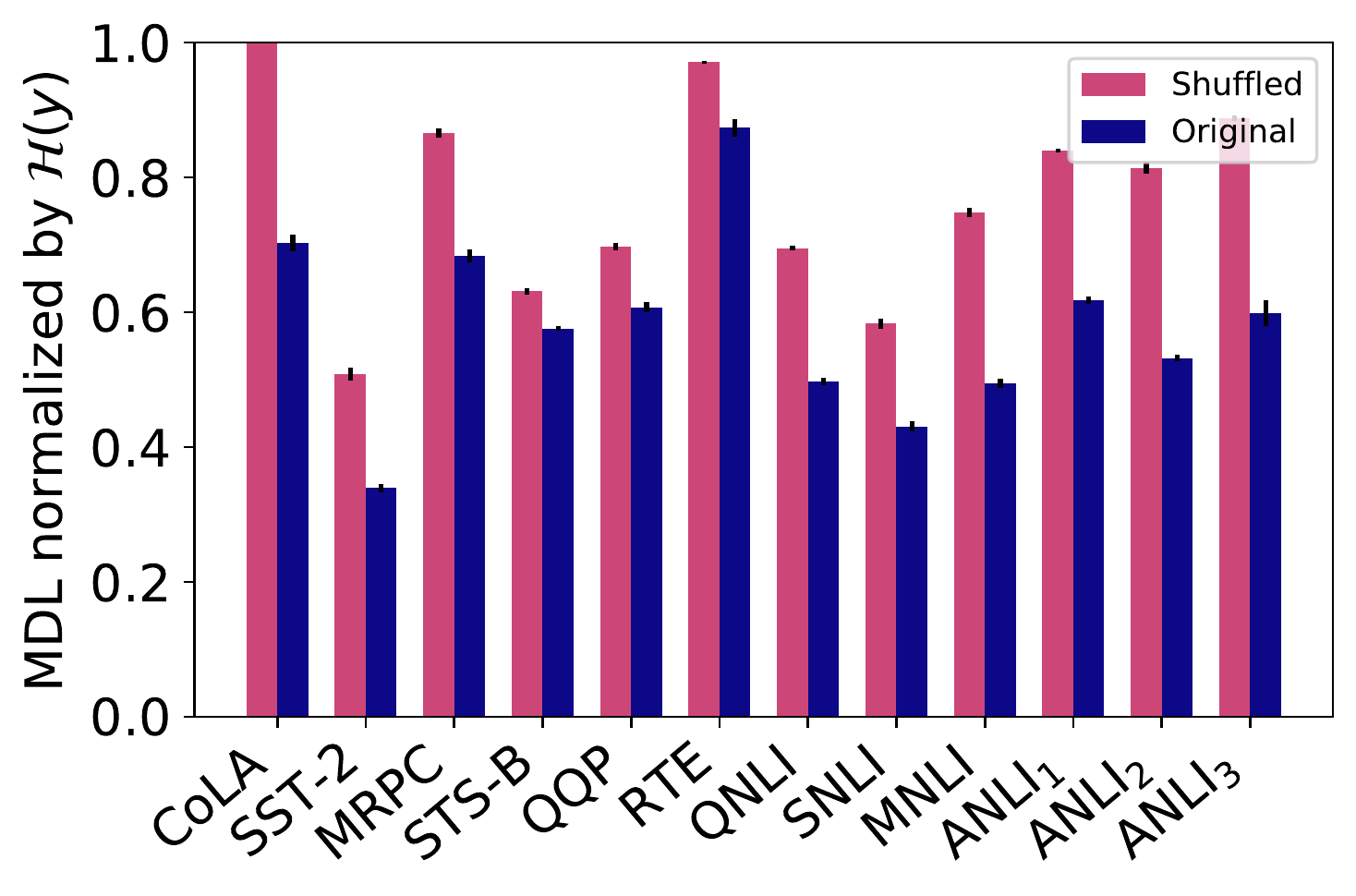}
    \caption{\textbf{\desc{} with/without Word Shuffling}, normalized by MDL when encoding labels with $p(y)$ for reference. Word order reduces \desc{} on all tasks.}
    \label{fig:glue-word-order}
\end{figure}

Recent work claims that state-of-the-art models do not use word
order for GLUE~\cite{pham2020order,sinha2020unnatural,gupta2021bert}, so we use RDA to examine the utility of word order on GLUE, by testing the value of the capability to unshuffle input words when they have been shuffled.

Fig.~\ref{fig:glue-word-order} shows \desc{} with and without shuffling, normalized by the \desc{} of the label-only prior $p(y)$ as a baseline.
Word order helps to obtain smaller \desc{} on all tasks.
For example, on MNLI, adding word order enables the labels to be compressed from $75\% \rightarrow 50\%$ of the baseline compression rate.
For CoLA, the linguistic acceptability task, input word order is necessary to compress labels at all.
Prior work may have come to different conclusions about the utility of word order because they evaluate the behavior of trained models on out-of-distribution (word-shuffled) text, while RDA estimates an intrinsic property of the dataset.

\section{Related Work}
\label{sec:Related Work}

In addition to prior work on data analysis (\S\ref{sec:Introduction}), there has been much work on model analysis~\citep[e.g.,][]{shi-etal-2016-string,alain2018understanding,conneau-etal-2018-cram,jia-liang-2017-adversarial}.
This line of work sometimes uses similar techniques, such as input replacement~\citep{perez-etal-2019-finding,jiang-bansal-2019-avoiding,pham2020order,sinha2020unnatural,gupta2021bert} and estimating description length~\citep{voita-titov-2020-information,whitney2020evaluating,lovering2021predicting} or other information-theoretic measures~\citep{pimentel-etal-2020-information}, but for a very different end: to understand how models behave and what their representations encode.
While model probing can uncover characteristics of the training data~\citep[e.g., race and gender bias;][]{caliskan2017semantics}, models also reflect other aspects of learning~\cite{zhao-etal-2017-men}, such as the optimization procedure, inductive bias of the model class and architecture, hyperparameters, and randomness during training.
Instead of indirectly examining a dataset by probing models, we directly estimate a property intrinsic to the dataset.
For further related work, see Appendix \S\ref{sec:Additional Related Work}.

\section{Conclusion}
In this work, we proposed Rissanen Data Analysis (RDA), a method for examining the characteristics of a dataset.
We began by viewing the labels of a dataset as being generated by a program over the inputs, then positing that a capability is helpful if it reduces the length of the shortest label-generating program.
Instead of evaluating minimum program length directly, we use block-wise prequential coding to upper bound Minimum Description Length (MDL).
While the choice of learning algorithm $\mathcal{A}$ influences absolute \desc{} values, we only interpret \desc{} \textit{relative} to other \desc{} values estimated with the same $\mathcal{A}$.
In particular, we conduct RDA by comparing \desc{} with or without access to a subroutine with a certain capability, and we say that a capability is useful when invoking the subroutine reduces \desc{}.

We then conducted an extensive empirical analyses of various datasets with RDA.
First, we showed that RDA provides intuitive results on a carefully-controlled synthetic task.
Next, we used RDA to evaluate the utility of generating and answering subquestions in answering a question, finding that subquestions are indeed useful.
For NLI, we found it helpful to include rationales and explanations.
Finally, we showcased the general nature of RDA by applying it on a variety of other NLP tasks, uncovering the value of word order across all tasks, as well as the most useful parts of speech for different tasks, among other things.
Our work opens up ample opportunity for future work: automatically uncovering dataset biases when writing data statements~\cite{gebru2018datasheets,bender-friedman-2018-data}, selecting the datasets to include in future benchmarks, discovering which capabilities are helpful for different tasks, and also expanding on RDA itself, e.g., by investigating the underlying data distribution rather than a particular dataset~\cite{whitney2020evaluating}.
Overall, RDA is a theoretically-justified tool that is empirically useful for examining the characteristics of a wide variety of datasets.


\section*{Acknowledgments}
We thank Paul Christiano, Sam Bowman, Tim Dettmers, Alex Warstadt, Will Huang, Richard Pang, Yian Zhang, Tiago Pimentel, and Patrick Lewis for helpful feedback.
We are grateful to OpenAI for providing access to GPT-3 via their API Academic Access Program.
We also thank Will Whitney and Peter Hase for helpful discussions, Adina Williams for the list of gendered words, Shenglong Wang for cluster support, and Amanda Ngo for figure design help.
\hotpot{} is licensed under CC BY-SA 4.0.
KC is partly supported by Samsung Advanced Institute of Technology (Next Generation Deep Learning: from pattern recognition to AI) and Samsung Research (Improving Deep Learning using Latent Structure).
KC also thanks Naver, eBay, NVIDIA, and NSF Award 1922658 for support.
EP is grateful to NSF and Open Philanthropy for fellowship support.

\bibliography{anthology_small,icml2021}
\bibliographystyle{acl_natbib}

\clearpage
\appendix

\section{Task Details}
\label{sec:Task Details}

\subsection{CLEVR}
\label{ssec:CLEVR Details}
We examine three question categories in CLEVR which have 1-2 relevant subquestions.
``Integer Comparison'' questions ask to compare the numbers of two kinds of objects and have two subquestions, i.e., ``Are there more cubes than spheres?'' where the two subquestions are ``\textit{How many cubes are there?}'' and ``\textit{How many spheres are there?}''
``Attribute Comparison'' questions ask to compare the properties of two objects, i.e., ``\textit{Is the metal object the same color as the rubber thing?}'', where there are two subquestions which each ask about the property of a single object, i.e., ``\textit{What color is the metal object?}'' and ``\textit{What color is the rubber thing?}''
``Same Property As'' questions ask whether or not one object has the same property as another object, i.e., ``\textit{What material is the sphere with the same color as the rubber cylinder?}'', where there is one subquestion that asks about a property of one object, i.e., ``\textit{What color is the rubber cylinder?}''
To obtain oracle subanswers, we use ground-truth programs given by CLEVR that can be executed over a symbolic, graph-based representation of the image to answer each question.
For each question category above, we evaluate the subprogram corresponding to its subquestion(s) to generate oracle subanswer(s).

\subsection{GLUE}
\label{ssec:GLUE Details}

GLUE consists of 9 tasks.
Two are single-sentence classification;
\textit{CoLA}~\citep[Corpus of Linguistic Acceptability;][]{warstadt2018neural} involves determining if a sentence is linguistically acceptable or not, while \textit{SST-2}~\citep[Stanford Sentiment Treebank 2;][]{socher-etal-2013-recursive} involves predicting if a sentence has positive or negative sentiment.
Three tasks involve determining if two sentences are similar or paraphrases of each other:
\textit{MRPC}~\citep[Microsoft Research Paragraph Corpus;][]{dolan-brockett-2005-automatically}, \textit{QQP} (Quora Question Pairs)\footnote{\href{data.quora.com/First-Quora-Dataset-Release-Question-Pairs}{data.quora.com/First-Quora-Dataset-Release-Question-Pairs}}, and
\textit{STS-B}~\citep[Semantic Textual Similarity Benchmark;][]{cer-etal-2017-semeval}.
The rest are NLI tasks: \textit{QNLI}~\citep[Question NLI, derived from SQuAD;][]{rajpurkar-etal-2016-squad}, \textit{RTE}~\citep[Recognizing Textual Entailment;][]{bentivogli2009fifth}, \textit{WNLI}~\citep[Winograd NLI;][]{levesque2012winograd}, and \textit{MNLI}~\citep[Multi-genre NLI;][]{williams-etal-2018-broad}.

\section{Model Training Details}
\label{sec:Training Details}

\begin{table*}[t]
\begin{center}
\begin{tabular}{lccccc}
\toprule
\bf Hyperparam  & \longformer & \roberta & \bart & \albert & \gpt \\
\midrule 
Learning Rate   & \tiny{\{3e-5, 5e-5, 1e-4\}} & \tiny{\{1e-5, 2e-5, 3e-5\}} & \tiny{\{5e-6, 1e-5, 2e-5\}} & \tiny{\{2e-5, 3e-5, 5e-5\}} & \tiny{\{6.25e-5, 3.125e-5, 1.25e-4\}} \\
Batch Size      & 32          &\{16, 32\}& \{32, 128\}& \{32, 128\}& 32 \\
Max Epochs      & 6           & 10       & 10         & 3          & 3 \\
Weight Decay    & 0.01        & 0.1      & 0.01       & 0.01       & 0.01 \\
Warmup Ratio    & 0.06        & 0.06     & 0.06       & 0.1        & 0.002 \\
Adam $\beta_2$  & 0.999       & 0.98     & 0.98       & 0.999      & 0.999 \\
Adam $\epsilon$ & 1e-6        & 1e-6     & 1e-8       & 1e-6       & 1e-8 \\
Grad. Clip Norm & $\infty$   & $\infty$ & $\infty$   & 1          & 1 \\
\bottomrule
\end{tabular}
\end{center}
\caption{
Training hyperparameters for all transformer models, based on those from each model's original paper. Column names refer to model types, including models of different sizes or trained from scratch with the same architecture.}
\label{tab:training-hyperparameters}
\end{table*}

\subsection{Distilled Language Model}
\label{ssec:Distilled Language Model Decompositions}

\subsubsection{Language Model Decompositions}
\label{sssec:Language Model Decompositions}
Large language models (LMs) are highly effective at text generation~\cite{brown2020language} but have not yet been explored in the context of question decomposition.
In particular, one obstacle is the sheer computational and monetary cost associated with such models.
We thus use an LM to generate question decompositions while conditioning on a few labeled question-decomposition pairs, and then we train a smaller, sequence-to-sequence model on the generated question-decomposition pairs, which we use to efficiently decompose many questions. Our approach, which we call Distilled Language Model (DLM), leverages the large LM to produce pseudo- training data for a more efficient model.

As our LM, we use the 175B parameter, pretrained GPT-3 model~\cite{brown2020language} via the OpenAI API.\footnote{\href{https://beta.openai.com/}{{https://beta.openai.com/}}}
We label the maximum number of question-decomposition pairs that fit in the context window of GPT-3 (2048 tokens or 46 question-decompositions).
For labeling, we sample questions randomly from \hotpot's training set.
To condition the LM, we format question-decomposition pairs as ``[Question] = [Decomposition]'', where the decomposition consists of several consecutive subquestions.
We concatenate the pairs, each on a new line, with a new question on the final line to form a prompt.
We then generate from the LM, conditioned on the prompt.
For decoding, we found that GPT-3 copies the question as the decomposition with greedy decoding.
Therefore, we use a sample-and-rank decoding strategy, to choose the best decoding out of several possible candidates.
We sample 16 decompositions with top-p sampling~\cite{holtzman2020curious} with $p=0.95$, rank decompositions from highest to lowest based their average token-level log probability, and choose the highest-ranked decomposition which satisfies the basic sanity checks for decomposition from~\citet{perez-etal-2020-unsupervised}.
The sanity checks avoid the question-copying failure mode by checking if a decomposition has (1) more than one subquestion (question mark),
(2) no subquestion which contains all words in the multi-hop question, and (3) no subquestion longer than the multi-hop question.
We generate decompositions for \hotpot{} dev questions, which we estimate costs $\$0.15$ per example or $\$1.1$k for the 7405 dev examples via the OpenAI API. Decomposing all 90447 training examples would roughly cost an extra $\$13.3$k, motivating distillation.

\subsubsection{Distilling Decompositions}
\label{sssec:Distilling Decompositions}
As our distilled, sequence-to-sequence model, we use the 3B parameter, pretrained T5 model~\cite{raffel2020exploring} via HuggingFace Transformers~\cite{wolf-etal-2020-transformers}.
We finetune T5 on our question-decomposition examples and then use it to generate subquestions for all training questions.

To finetune T5, we split our question-decomposition examples into train (80\%), dev (10\%), and test (10\%) splits.
We finetune T5 with a learning rate of $1e-4$, and we sweep over label smoothing $\in \{0.1, 0.2, 0.4, 0.6\}$, number of training epochs $\in \{3, 5, 10\}$, and batch size in $\in \{16, 32, 64\}$, choosing the best hyperparmeters ($0.1, 3, 64$, respectively) based on dev BLEU~\cite{papineni-etal-2002-bleu}.
We stop training early when dev BLEU does not increase after one training epoch.
We generate decompositions using beam search of size $4$ and length penalty of $0.6$ as in~\citet{raffel2020exploring}, achieving a test BLEU of 50.7.
We then finetune a new T5 model using the best hyperparameters on all question-decomposition examples except for a small set of 200 examples used for early stopping.

\subsection{\longformer}
\label{ssec:Longformer}
Similar to~\citet{beltagy2020longformer}, we train \longformer{} models for up to $6$ epochs, stopping training early if dev loss doesn't decrease after one epoch. We sweep over learning rate $\in \{3\times10^{-5}, 5\times10^{-5}, 1\times10^{-4}\}$.

\subsection{Regression}
\label{ssec:Regression}
STS-B is a regression task in GLUE where labels are continuous values in $[0, 5]$. Here, we learn to minimize mean-squared error, which is equivalent to minimizing log-likelihood and thus codelength.\footnote{\citet{cover2006elements} justifies the relationship between log-likelihood and codelength for continuous values.} We treat each scalar prediction as the mean of a Gaussian distribution and tune a single standard deviation parameter shared across all predictions from a single model. We choose the variance based on dev log-likelihood using grid search over $[10^{-2.5}, 10^{1.5}]$ with 1000 log-uniformly spaced samples.
To send the first block of labels, Alice and Bob use a uniform distribution over $[0, 5]$.

The FastText library only supports classification, so we convert STS-B to 26-way classification by rounding label values to the nearest 0.2, following~\citet{raffel2020exploring}. We compute a mean prediction by evaluating the average class label value when marginalizing over class probabilities. We then tune variance on dev as usual.

\begin{figure*}[t]
	\centering
    \begin{subfigure}
        \centering
        \includegraphics[scale=.36]{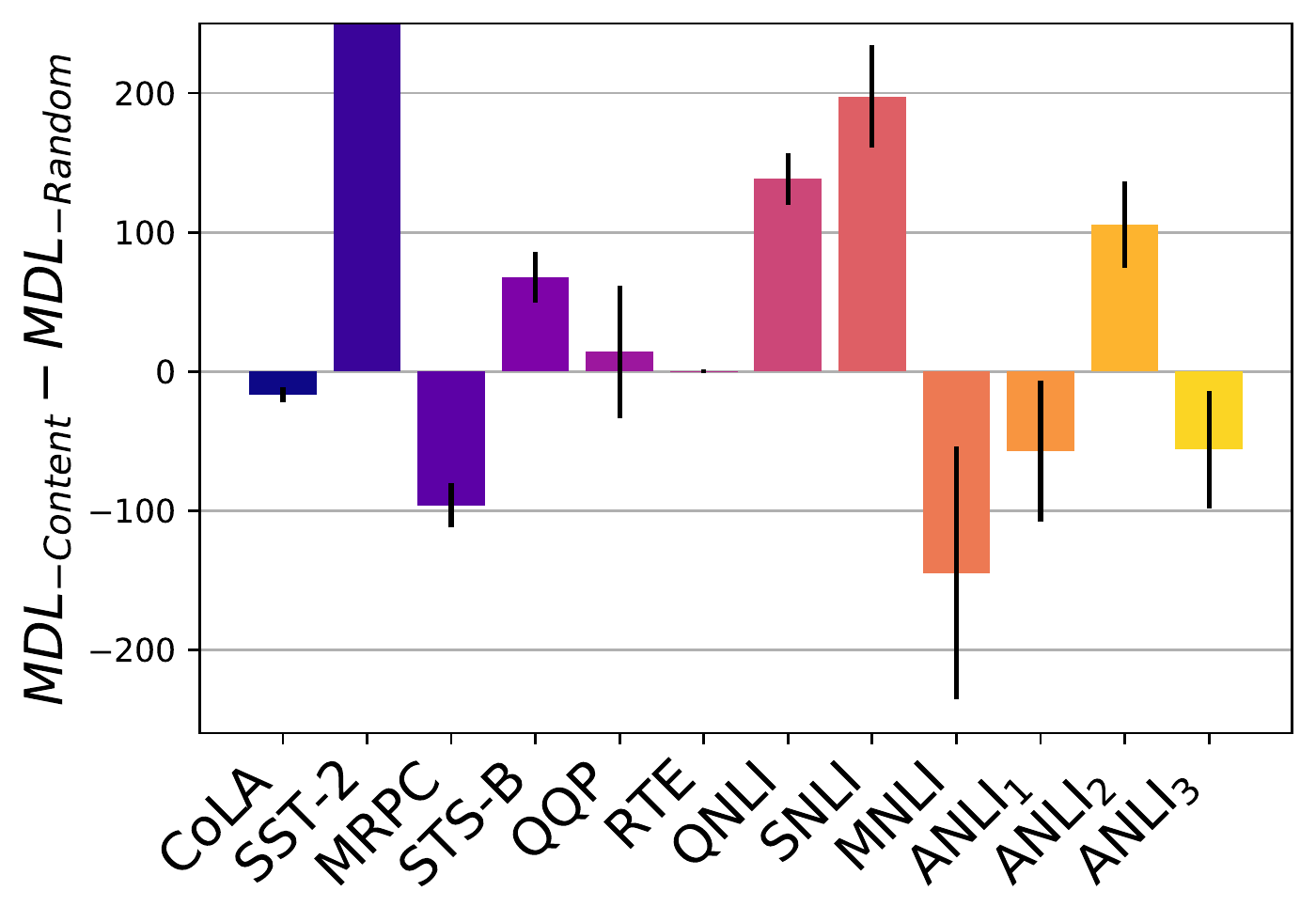}
    \end{subfigure}
    \begin{subfigure}
		\centering
        \includegraphics[scale=.36]{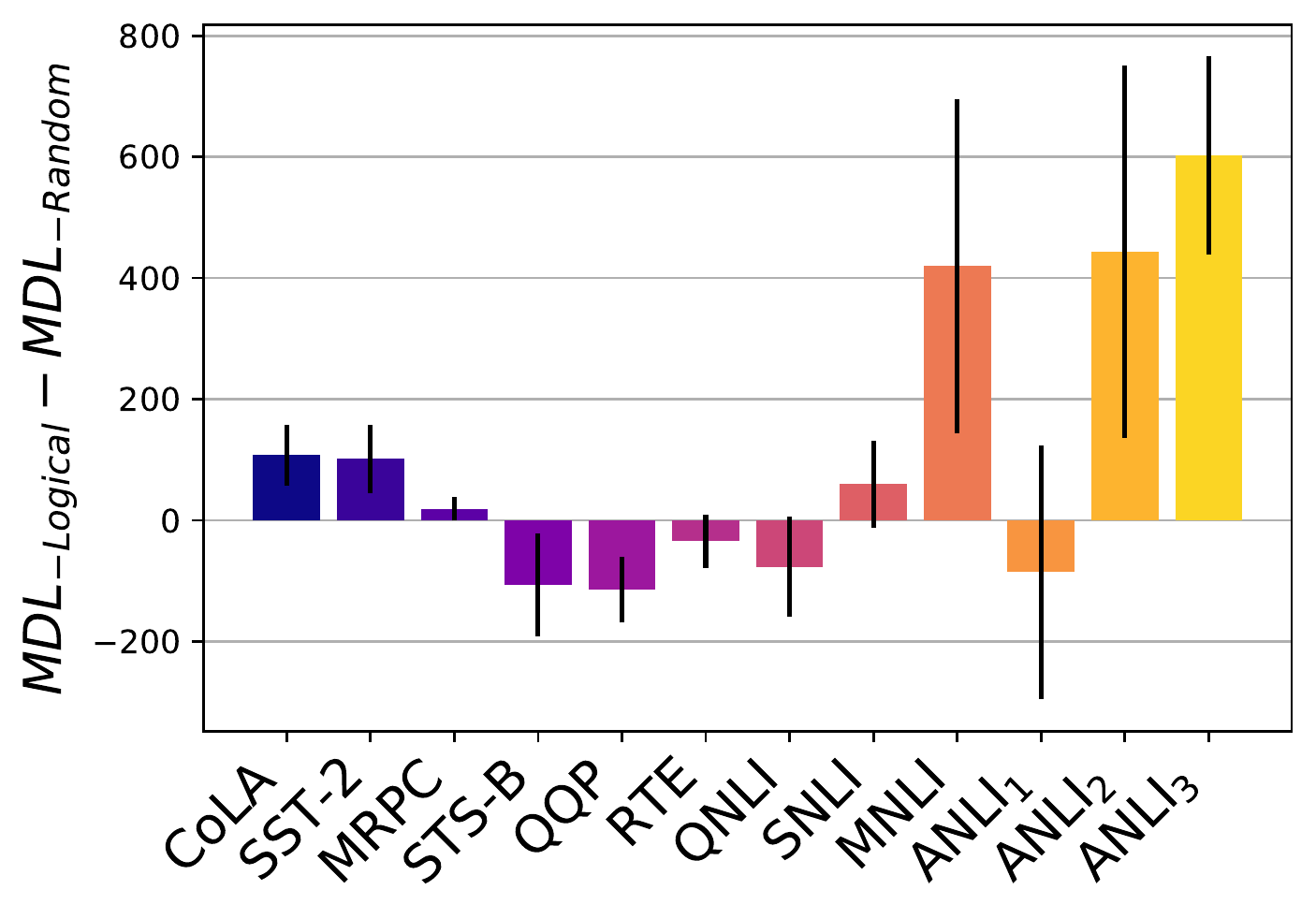}
    \end{subfigure}
    \begin{subfigure}
		\centering
        \includegraphics[scale=.36]{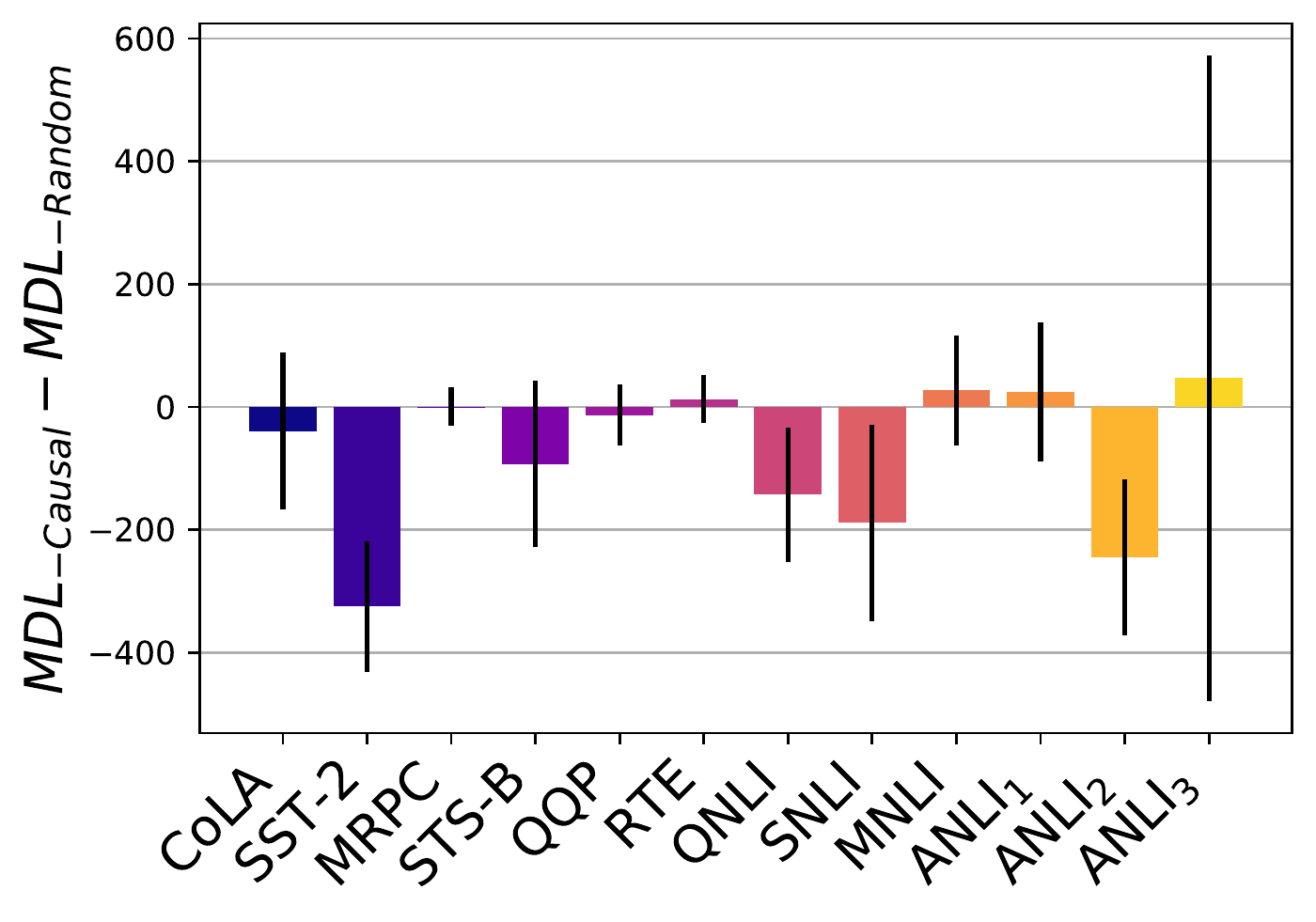}
    \end{subfigure}
    \caption{Difference between \desc{} when we mask input words that are (1) of a given type and (2) randomly chosen with the same frequency as (1). Mean and std. err. over 5 random seeds for content words (\textbf{left}), logical words (\textbf{middle}), and causal words (\textbf{right}).}
    \label{fig:glue-other-word-types}
\end{figure*}

\label{ssec:Is it helpful to answer subquestions? Details}
\begin{figure}[t]
	\centering
    \includegraphics[scale=0.46]{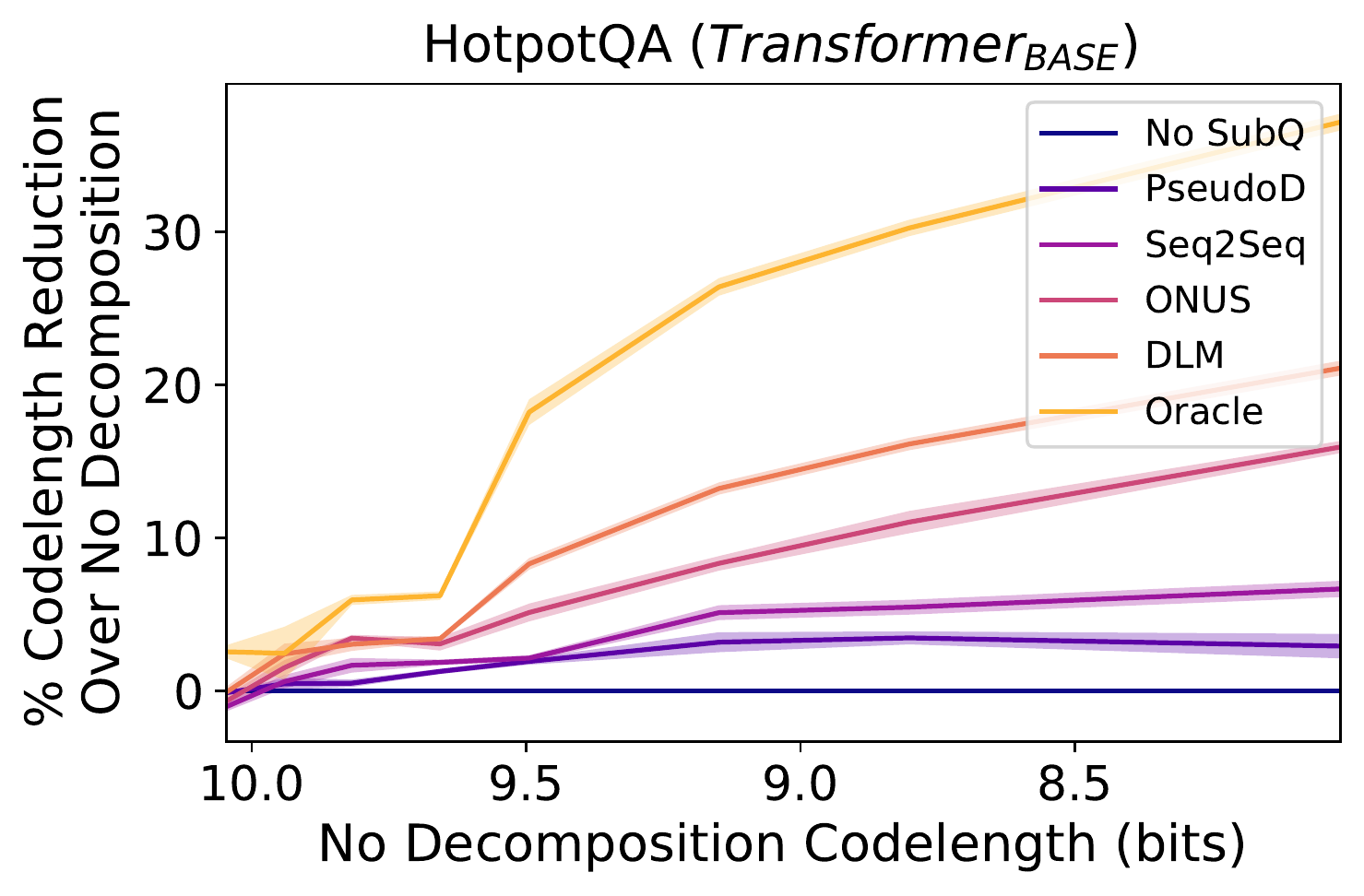}
    \caption{The reduction in codelength over the no-decomposition baseline from using subanswers from various decomposition methods (mean and std. error for \transformerbase).}
    \label{fig:hotpot-codelength-reduction-transformer-base}
\end{figure}

\subsection{Ensemble Model}
\label{ssec:Ensemble Model}

\subsubsection{FastText}
For the FastText classifier, we initialize with the 2M pretrained, 300-dimensional word vectors trained on Common Crawl (600B tokens).\footnote{\href{https://fasttext.cc/docs/en/english-vectors.html}{https://fasttext.cc/docs/en/english-vectors.html}}
We tune hyperparameters using the official implementation of automatic hyperparameter tuning, which we run for 2 hours, which is generally sufficient for $20+$ hyperparmeter trials and convergence on dev accuracy.
The tuning implementation chooses the hyperparameters based on dev accuracy instead of loss as we typically do, but our procedure of tuning a softmax temperature parameter helps FastText reach significantly below-baseline loss.

\subsubsection{Transformer Models}
The remaining models in our ensemble are transformer-based models trained with HuggingFace Transformers~\cite{wolf-etal-2020-transformers}.
Table~\ref{tab:training-hyperparameters} shows the hyperparameter ranges used for each model, which we chose based on those used in each model's original paper for GLUE.
For \transformer{} models trained from scratch on e-SNLI and GLUE, we use the \robertabase{} and \robertalarge{} architecture with the \roberta hyperparameters, except that we use a larger batch size ($\in \{64, 128\}$) for the \textsc{LARGE} transformer, which gave better results.

\subsection{Training Hardware}
We train FastText on 1 CPU core (40GB of memory) and FiLM on GeForce GTX 1080ti 11GB GPUs (10GB CPU memory).
We train other models with ``almost floating point 16'' mixed precision~\cite{micikevicius2018mixed} on 1 RTX-8000 48GB GPU (CPU memory of 100GB for \hotpot{} and 30GB for e-SNLI and GLUE).

\section{Additional Experiments}
\label{sec:Additional Experiments}

\subsection{Is it helpful to answer subquestions?}

In \S\ref{ssec:hotpot}, we found that decompositions increase in usefulness as the original, no-decomposition model's loss decreases, up until some point after which decompositions decrease in usefulness.
To examine if the same trend holds for other models, we show the same plot for \transformerbase{} in Fig.~\ref{fig:hotpot-codelength-reduction-transformer-base}.
Here, decompositions increase in usefulness as the codelength (loss) of the original model decreases.
However, for most decomposition methods, we do not find a point at which decompositions begin to decrease in usefulness, which we fits our hypothesis that the model must have enough data to learn the task directly in order for decompositions to become less useful.
However, the slope of improvement decreases (i.e., the second derivative is negative), suggesting that, given more training data, decompositions will also peak in usefulness for \transformerbase.

\subsection{Examining Text Datasets}
\label{ssec:Examining Text Datasets Details}

\paragraph{How useful are content words?}
\citet{sugawara2020assessing} hypothesized that ``content'' words are particularly useful for NLP tasks, taking content words to be nouns, verbs, adjectives, adverbs, or numbers.
We test their utility on GLUE, SNLI, and ANLI using RDA, by evaluating $\desc_{-\text{Content}}-\desc_{-\text{Random}}$ (Fig.~\ref{fig:glue-other-word-types} left).
The value is positive for SST-2, STS-B, QNLI, SNLI, and ANLI$_2$ and negative for MRPC, MNLI, ANLI$_1$, and ANLI$_3$.
In particular, the value for SST-2 is very high (1732), indicating that content words are important for sentiment classification, likely due to the importance of adjectives as found in \S\ref{ssec:pos}.
For QNLI, content words are important, despite earlier findings that each individual POS group (nouns, verbs, adjectives, or adverbs) was not important for QNLI (\S\ref{ssec:pos} Fig.~\ref{fig:glue-pos-ablations}), indicating that QNLI requires reasoning over multiple POS in tandem.

\paragraph{How useful are ``logical'' words?}
\citet{sugawara2020assessing} hypothesized that words that have to do with the logical meaning of a sentence (e.g., quantifiers and logical connectives) are useful for NLP tasks.
Using GLUE, SNLI, and ANLI, we test the usefulness of logical words, which we take as: \textit{all, any, each, every, few, if, more, most, no, nor, not, n't, other, same, some,} and \textit{than}~\citep[following][]{sugawara2020assessing}.
As shown in Fig.~\ref{fig:glue-other-word-types} (middle), $\desc_{-\text{Logical}}-\desc_{-\text{Random}}$ is positive for CoLA, SST-2, MNLI, ANLI$_2$, and ANLI$_3$ and negative for STS-B and QQP.
Notably, $\desc_{-\text{Logical}}-\desc_{-\text{Random}}$ is large for MNLI, ANLI$_2$, and ANLI$_3$, three entailment detection tasks, where we expect logical words to be important.

\paragraph{How useful are causal words?}
Another group of words that \citet{sugawara2020assessing} hypothesized are useful are words that express causal relationships: \textit{as, because, cause, reason, since, therefore,} and \textit{why}.
As shown in Fig.~\ref{fig:glue-other-word-types} (right), $\desc_{-\text{Causal}}-\desc_{-\text{Random}}$ to be within std. error of 0 for all tasks except SST-2, QNLI, SNLI, and ANLI$_2$, where $\desc_{-\text{Causal}}-\desc_{-\text{Random}} < 0$.
Thus, causal words do not appear particularly useful for GLUE.

\paragraph{Do datasets suffer from gender bias, even when controlling for word frequency?}

\begin{figure}[t!]
	\centering
    \includegraphics[scale=0.5]{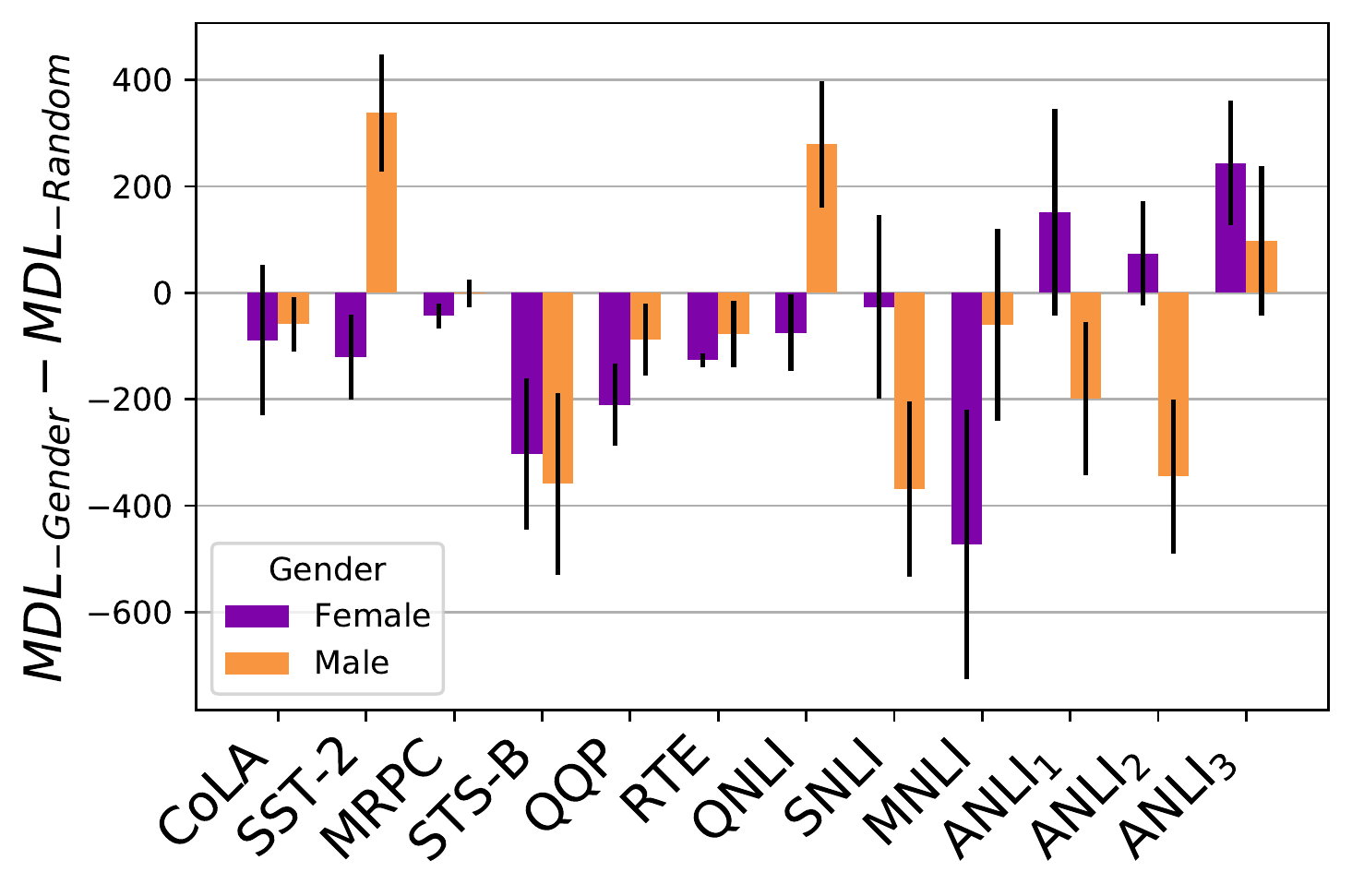}
    \caption{The difference between \desc{} when (1) masculine/feminine words are masked and (2) the same fraction of input words are masked at random.
    }
    \label{fig:gender-controlled}
\end{figure}

In \S\ref{ssec:gender-bias}, we assessed if datasets rely more on male- or female- gendered words by comparing \desc{} when masculine vs. feminine input words are masked, looking at $\desc_{-\text{Male}}-\desc_{-\text{Female}}$.
However, we may wish to focus on gender bias present in datasets beyond easy-to-detect differences in male- and female- gendered word frequency.
To control for frequency, we evaluate $\desc_{-\text{Male}}-\desc_{-\text{Random}}$ and $\desc_{-\text{Female}}-\desc_{-\text{Random}}$, as we did for our word type experiments.
We show results in Fig.~\ref{fig:gender-controlled}.
For SST-2 and QNLI, masculine words are more useful than randomly-chosen words, while feminine words are less useful than randomly-chosen words, a sign of gender bias.
Most tasks, however, do not show similar patterns of bias as SST-2 and QNLI do.

\paragraph{How useful is input length?}

\begin{figure}[t]
	\centering
    \includegraphics[scale=0.5]{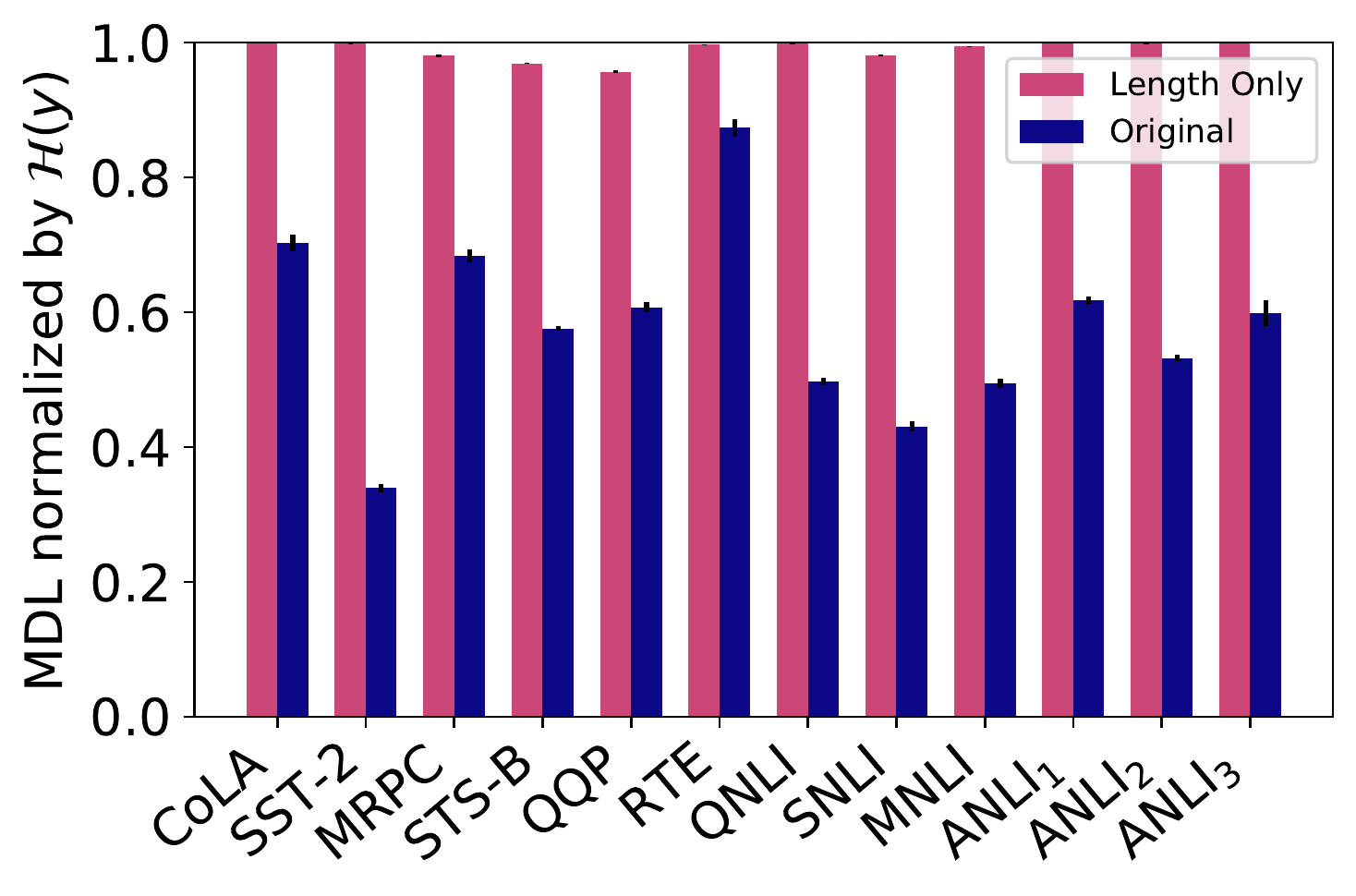}
    \caption{\textbf{\desc{} with length-only input} compared to \desc{} with original input, normalized by the MDL when encoding labels with $p(y)$ for reference. Length information reduces \desc{} over $p(y)$ on MRPC, STS-B, QQP, and SNLI, though not significantly.}
    \label{fig:length}
\end{figure}

Input text length can be highly predictive of the class label~\citep[see, e.g.,][]{dixon2018measuring}.
RDA can be used to evaluate text datasets for such length bias.
We evaluate \desc{} when only providing the input length, in terms of number of tokens (counted via spaCy).\footnote{Masking all input tokens gave similar results.}
As shown in Fig.~\ref{fig:length}, the labels in MRPC, STS-B, QQP, and SNLI can be compressed using the input length, though not to a large extent.
Other tasks cannot be compressed using length alone.
Our results on SNLI agree with~\citet{gururangan-etal-2018-annotation} who found that hypotheses were generally shorter for entailment examples and longer for neutral examples.
Similarly, they also found that length is less discriminative on MNLI compared to SNLI.

\section{Additional Related Work}
\label{sec:Additional Related Work}

Recent work has raised increasing awareness of the importance of characterizing the datasets we release, via datasheets~\cite{gebru2018datasheets} or data statements~\cite{bender-friedman-2018-data}.
A key motivation for datasheets is to inform machine learning practitioners of (1) biases that models may learn when trained on the data or (2) model weaknesses that may not be caught by testing on biased data.
As we saw earlier, RDA is a useful tool for catching such biases (e.g., gender bias) and thus for writing datasheets.

RDA shares high-level motivation with other data analysis methods that aim to measure intrinsic properties of the data.
For example, on NLI,~\citet{gururangan-etal-2018-annotation} measure the point-wise mutual information (PMI) between the label and occurrence of different keywords to find heuristics for SNLI.
Similarly,~\citet{rudinger-etal-2017-social} measure PMI between premise words and hypothesis words to uncover race, age, and gender stereotypes in crowdsourced SNLI hypotheses.
On MNLI,~\citet{mccoy-etal-2019-right} measure the accuracy of heuristics such as ``Assume that a premise entails all hypotheses constructed from words in the premise.''
These methods capture the relationship between output labels and an input feature, considered \textit{in isolation}, but some features may only be useful when provided along with other features.
In such cases, RDA can still capture the utility of the feature.

Other work aims to analyze properties of individual examples in a dataset.
For instance, \citet{koh2017understanding} use influence functions to determine which training instances are most responsible for a particular test-time prediction, e.g., for image classification.
\citet{brunet2019understanding} use influence functions to find the training documents most responsible for producing gender-biased word embeddings.
Item response theory~\cite{baker2004item} has been used to find the most challenging examples for current models~\cite{hopkins-may-2013-models,lalor-etal-2019-learning,martinez2019item}.
Instead of examining individual examples, we examine general characteristics of the dataset as a whole.




\end{document}